\documentclass{article} %
\usepackage{iclr2027_conference,times}
\iclrfinalcopy

\usepackage{appendix}
\usepackage{enumitem}
\setlist{topsep=0pt, leftmargin=*}

\usepackage{algorithm}
\usepackage{algpseudocode}

\usepackage{bm}
\usepackage{bbm}
\usepackage{amsmath}
\usepackage{amssymb}
\usepackage{mathtools}
\usepackage{amsthm}

\theoremstyle{plain}
\newtheorem{theorem}{Theorem}[section]

\theoremstyle{definition}

\theoremstyle{remark}

\newtheoremstyle{named}{}{}{\itshape}{}{\bfseries}{.}{.5em}{\thmnote{#3's }#1}
\theoremstyle{named}

\theoremstyle{plain} %
\newcommand{\thistheoremname}{}
\newtheorem{genericthm}[theorem]{\thistheoremname}

\newtheorem*{genericthm*}{\thistheoremname}
\newenvironment{namedthm*}[1]
  {\renewcommand{\thistheoremname}{#1}%
   \begin{genericthm*}}
  {\end{genericthm*}}

\newcounter{numrel}%
\counterwithin*{numrel}{section}%

\newcommand{\abs}[1]{\left\lvert#1\right\rvert }

\DeclareMathOperator*{\argmin}{arg\,min}

\newcommand{\R}{\mathbb{R}}

\newcommand{\bI}{\mathbf{I}}

\newcommand{\bx}{\bm{x}}
\newcommand{\by}{\bm{y}}

\newcommand{\bw}{\bm{w}}

\newcommand{\btheta}{\bm{\theta}}

\usepackage{hyperref}
\usepackage{url}

\usepackage[utf8]{inputenc} %
\usepackage[T1]{fontenc}    %
\usepackage{xurl}           %
\usepackage{booktabs}       %
\usepackage{amsfonts}       %
\usepackage{nicefrac}       %
\usepackage{microtype}      %
\usepackage{xcolor}         %
\usepackage{graphicx}
\usepackage{array}
\usepackage{colortbl}       %
\usepackage{tabularx}
\usepackage{multirow}
\usepackage{caption}
\usepackage{subcaption}
\usepackage{wrapfig}
\usepackage{placeins}
\usepackage{tikz}
\usetikzlibrary{arrows.meta,positioning}

\usepackage[capitalize,noabbrev]{cleveref}

\newlength{\radSavedTextFloatSep}
\newlength{\radSavedFloatSep}
\newlength{\radSavedInTextSep}
\usepackage[most]{tcolorbox}
\definecolor{contribframe}{HTML}{8AAAD1}
\definecolor{contribbg}{HTML}{F5F8FC}
\newtcolorbox{mainbox}[1][]{%
  colframe=contribframe,
  colback=contribbg,
  boxrule=0.6pt,
  arc=2pt,
  breakable,
  before upper={\textbf{Main Contributions.}\par\smallskip},
  #1
}

\newcommand{\method}{RAD}
\newcommand{\meanstd}[2]{\ensuremath{#1_{\pm #2}}}
\newcommand{\bestmeanstd}[2]{%
  \ensuremath{\mathbf{#1}_{\boldsymbol{\pm}\mathbf{#2}}}}
\newcommand{\secondmeanstd}[2]{%
  \ensuremath{\underline{#1_{\pm #2}}}}
\newcommand{\baselinegainlabel}{\(\Delta\) vs.\ best baseline}
\newcommand{\baselinegain}[1]{\ensuremath{(#1)}}
\newcolumntype{C}[1]{>{\centering\arraybackslash}p{#1}}
\definecolor{radblue}{HTML}{2A78D6}
\definecolor{raddark}{HTML}{1B3A6B}
\definecolor{radteal}{HTML}{15976B}
\definecolor{radamber}{HTML}{EDA100}
\definecolor{radmuted}{HTML}{6B6A64}

\definecolor{radred}{HTML}{C8352E}
\definecolor{radrowbg}{HTML}{FBECEA}
\definecolor{radgaintext}{HTML}{1B7A3A}
\definecolor{radheadbg}{HTML}{EFEFF5}
\definecolor{radink}{HTML}{1F2937}
\definecolor{radinksecondary}{HTML}{4B5563}
\definecolor{radneutralbg}{HTML}{F3F4F6}
\newcommand{\radrow}{\rowcolor{radrowbg}}
\newcommand{\radgainrow}{}
\newcommand{\radhead}{\rowcolor{radheadbg}}
\newcommand{\radlabel}[1]{\textcolor{radred}{\textbf{#1}}}
\newcommand{\radgain}[1]{\textcolor{radgaintext}{\baselinegain{#1}}}

\newcommand{\methodoverviewfigure}{%
\begin{figure}[t]
    \centering
    \resizebox{\linewidth}{!}{%
    \begin{tikzpicture}[
        >=Latex,
        state/.style={
            draw=radinksecondary,
            fill=radneutralbg,
            circle,
            minimum size=1.05cm,
            align=center,
            font=\small
        },
        stage/.style={
            draw=radblue!70,
            fill=radblue!6,
            rounded corners=2pt,
            align=center,
            text width=3.45cm,
            minimum height=1.28cm,
            inner sep=5pt,
            font=\small
        },
        init/.style={
            draw=radred,
            fill=radred!6,
            rounded corners=2pt,
            align=center,
            text width=2.75cm,
            minimum height=1.28cm,
            inner sep=5pt,
            font=\small
        },
        weightbox/.style={
            draw=radred,
            fill=radred!6,
            rounded corners=2pt,
            align=center,
            text width=6.45cm,
            minimum height=2.00cm,
            inner sep=6pt,
            font=\small
        },
        actual/.style={
            draw=radinksecondary,
            fill=radneutralbg,
            rounded corners=2pt,
            align=center,
            text width=5.55cm,
            minimum height=2.00cm,
            inner sep=6pt,
            font=\small
        },
        flow/.style={->,thick,draw=radinksecondary}
    ]
        \node[state] (theta) {$\btheta$};
        \node[init,right=0.80cm of theta] (initialize) {
            \textbf{\color{radred}1. Initialize}\\[2pt]
            $\displaystyle w_t^{(0)}=c,\ \forall t$
        };
        \node[stage,right=0.45cm of initialize] (lookahead) {
            \textbf{\color{radblue!70!black}2. Temporary lookahead}\\[2pt]
            $\displaystyle
            \btheta^+
            =\btheta-\eta\nabla_{\btheta}L_D(\btheta,c\mathbf 1)$
        };
        \node[stage,right=0.50cm of lookahead] (reward) {
            \textbf{\color{radblue!70!black}3. Post-update reward}\\[2pt]
            $\displaystyle J_R(\btheta^+)$
        };

        \node[weightbox,below=0.35cm of reward] (weights) {
            \textbf{\color{radred}4. Update token weights}\\[2pt]
            $\displaystyle
            \begin{aligned}
                I_t(\by)
                &=-\eta\left\langle
                \nabla_{\btheta^+}J_R(\btheta^+),
                \nabla_{\btheta}\ell_{D,t}(\btheta;\by)\right\rangle,\\[3pt]
                w_t(\by)&=\Pi_{[c-1,c+1]}\!\left(c+\gamma I_t(\by)\right)
            \end{aligned}
            $
        };

        \node[actual,left=1.10cm of weights] (student) {
            \textbf{\color{radink}5. Commit weighted update}\\[2pt]
            $\displaystyle
            \btheta\gets\btheta-\eta\nabla_{\btheta}L_D(\btheta,\bw)$
        };

        \draw[flow] (theta) -- (initialize);
        \draw[flow] (initialize) -- (lookahead);
        \draw[flow] (lookahead) -- (reward);
        \draw[flow] (reward.south) --
            node[midway,right=2pt,font=\scriptsize] {differentiate}
            (weights.north);
        \draw[flow] (weights) --
            node[midway,above=2pt,font=\scriptsize] {$\bw$} (student);
        \draw[->,thick,dashed,draw=radinksecondary]
            (student.west) -- ++(-0.50,0) -|
            node[pos=0.30,below=2pt,font=\scriptsize] {next iteration}
            (theta.south);
    \end{tikzpicture}%
    }
    \caption{\textbf{One iteration of reward-aligned distillation (RAD),
    illustrated with SGD.}
    (1) Starting from student \(\btheta\), assign the initial token weight \(c\) to every token.
    (2) Take one temporary step on the weighted distillation objective
    to obtain \(\btheta^+\), as in \cref{eq:lookahead}.
    (3) Sample separate meta rollouts from \(\btheta^+\) to estimate the post-update
    reward gradient.
    (4) Differentiate this reward through the temporary step to score each
    token using \cref{eq:token-score}. 
    Positive scores increase token weights and negative
    scores decrease them; projected ascent clips the updated weights to
    \([c-1,c+1]\), as in \cref{eq:projected-weight-update}.
    (5) Discard the temporary update and take one weighted distillation step
    from the original \(\btheta\). 
    Reset the weights to \(c\) for the next batch.}
    \label{fig:rad-overview}
\end{figure}
}

\title{Reward-Aligned Distillation}

\author{%
  \begin{tabular}{cccc}
    \textbf{Yilan Chen}\textsuperscript{1,}\setcounter{footnote}{1}\thanks{Work done during an internship at AWS AI Labs. Yilan Chen: \href{mailto:yic031@ucsd.edu}{\texttt{yic031@ucsd.edu}}; Zian Su: \href{mailto:su284@purdue.edu}{\texttt{su284@purdue.edu}}.} &
    \textbf{Zian Su}\textsuperscript{2,$\dagger$} &
    \textbf{Manish Prajapat}\textsuperscript{3} &
    \textbf{Dingmin Wang}\textsuperscript{3}
    \\[0.4em]
    \multicolumn{4}{c}{%
      \textbf{Rui Song}\textsuperscript{4}
      \qquad\qquad
      \textbf{Rein Houthooft}\textsuperscript{3}
      \qquad\qquad
      \textbf{Vijay Lingam}\textsuperscript{3,}\setcounter{footnote}{0}\thanks{Corresponding author: Vijay Lingam, \href{mailto:vlingam@amazon.com}{\texttt{vlingam@amazon.com}}.}
    }
  \end{tabular}
  \\[1.2em]
  \textsuperscript{1}UC San Diego
  \qquad
  \textsuperscript{2}Purdue University
  \qquad
  \textsuperscript{3}AWS AI Labs
  \qquad
  \textsuperscript{4}Amazon
}

\begin{document}

\maketitle
\lhead{Preprint}

\begin{abstract}
Reinforcement learning with verifiable rewards improves the reasoning capabilities of large language models, but remains computationally expensive and sample inefficient. On-policy distillation (OPD) provides dense token-level supervision from a stronger teacher, yet standard OPD distills every token regardless of whether the teacher's preference actually improves task reward. Existing token-selection methods rely on proxies such as uncertainty or teacher--student divergence, which need not identify reward-improving supervision. We introduce Reward-Aligned Distillation (RAD), a bilevel framework that weights distillation signals based on their predicted effect on the student's post-update reward. RAD differentiates through a temporary optimizer step to score each token's supervision, then updates the token weights by projected ascent from the initial token weights. The resulting weights can strengthen,
suppress, or reverse individual distillation-gradient contributions. At a matched training-response budget, RAD improves mean accuracy on six competition-math benchmarks over the strongest baselines by 6.5 percentage points for Qwen3-1.7B (non-thinking) and 2.5 points for Qwen3-4B-Instruct-2507. RAD-Simple, a simplified variant, achieves comparable math accuracy at the same training-response budget. On LiveCodeBench-v6, RAD reaches 54.69\% and 58.67\%
pass@1 with Qwen3-1.7B (non-thinking) and OLMo3-7B-Instruct, respectively,
exceeding the strongest baselines by 8.7 and 6.4 points. These results support reward-guided weighting of teacher supervision across model families and tasks.
\looseness=-1

\end{abstract}

\begin{figure}[H]
\centering
\includegraphics[width=0.95\linewidth]{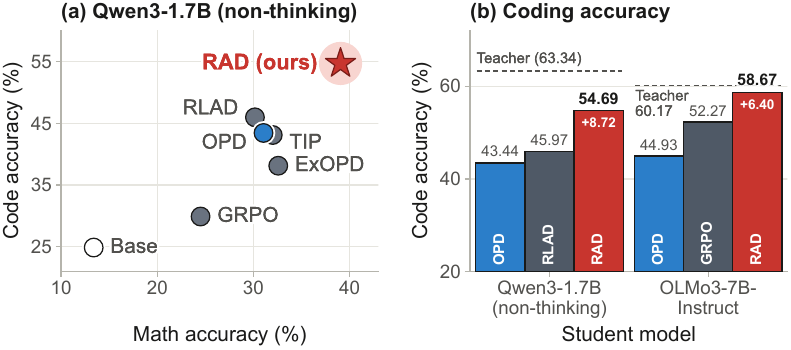}
\caption{Reward-aligned distillation (RAD) consistently outperforms OPD and other baselines
on mathematical reasoning and
code generation tasks. The teachers are Qwen3-30B-A3B-Instruct-2507 for
the Qwen student and OLMo3.1-32B-Instruct for the OLMo student.
(a) Rollout-matched Qwen3-1.7B (non-thinking)
results on both math and coding tasks
(\cref{tab:main-results,tab:coding-results}).
(b) LiveCodeBench-v6 pass@1 for Qwen3-1.7B (non-thinking) and
OLMo3-7B-Instruct, comparing RAD at step 100 with baselines at step 200.
The grey bars show the strongest baselines, RLAD and GRPO,
respectively; each RAD bar gives its gain over that baseline (pp).
Dashed lines mark the corresponding teachers.}
\label{fig:headline-results}
\end{figure}

\section{Introduction}

Reinforcement learning with verifiable rewards (RLVR) has been shown effective for improving
mathematical reasoning \citep{IMO2025} and code generation \citep{jaech2024openai, ekbote2025murphy} abilities,
but its supervision is sparse: a long response often
receives only a single sequence-level reward
\citep{shao2024deepseekmath,guo2025deepseek}. On-policy distillation offers a
complementary signal by evaluating student-generated responses under a
strong teacher and supplying dense token-level guidance
\citep{agarwal2024policy, gu2024minillm}. This approach is
appealing because the student continues to explore on policy while learning
from the teacher at every generated token.

Standard OPD, however, implicitly treats the teacher as the target policy and
every teacher preference as worth imitating. This would be justified if the
teacher were optimal for the target reward, but a stronger teacher need not be
reward-optimal. Teacher--student differences may reflect style, formatting,
verbosity, or reasoning choices that do not improve task reward \citep{armandpour2026unmasking}. Consequently,
reducing teacher--student divergence does not guarantee reward improvement:
uniform distillation can spend capacity matching reward-irrelevant behavior or
move the student away from reward-improving solutions~\citep{xu2026tip,rgopd}.

The challenge is therefore to determine which teacher signals will benefit the student’s task performance. Token-selection methods~\citep{xu2026tip,xiao2026finding} use heuristic criteria such as uncertainty, teacher--student disagreement, and reasoning structure to selectively apply teacher supervision. These criteria identify potentially useful tokens, but do not directly assess whether the corresponding distillation update improves the student policy's task reward. We therefore seek to choose each token’s distillation weight based on its predicted effect on reward: \emph{retain teacher signals that help the student solve the task, and suppress or reverse those that hinder it.}

To this end, we introduce reward-aligned distillation (RAD), which weights teacher supervision by its predicted effect on task reward. Token weights affect reward indirectly, by controlling how the student learns from teacher supervision. RAD captures this dependency with a bilevel objective: the inner problem trains the student with weighted distillation, while the outer problem selects weights to maximize the trained student's reward (\autoref{eq:bilevel}). We approximate the inner optimization with one temporary optimizer step using token weights initialized to a tunable value. Differentiating the temporary student's reward with respect to these weights yields look-ahead scores that guide a projected ascent update. The resulting signed weights can strengthen, suppress, or reverse individual distillation-gradient contributions.\looseness=-1

\begin{mainbox}
\begin{itemize}
    \item We introduce Reward-Aligned Distillation (RAD), a bilevel framework that assigns 
    bounded, signed token weights according to their predicted effect on task reward (\cref{sec:bilevel}). 
    A one-step approximation yields optimizer-aware scores computed jointly through a single Jacobian--vector product (\cref{sec:general_rad}). 
    We also introduce RAD-Simple, which simplifies token scoring by removing the lookahead update and reusing distillation rollouts (\cref{sec:rad-lite}).

    \item At matched training-response budgets, RAD and RAD-Simple improve mean accuracy across six math benchmarks over the
    strongest baselines by up to \textbf{6.48} and \textbf{3.55} percentage points for Qwen3-1.7B (non-thinking) and
    Qwen3-4B-Instruct-2507, respectively. 
    On LiveCodeBench-v6, the corresponding pass@1 gains for RAD are \textbf{8.7} and \textbf{6.4} points for Qwen3-1.7B and OLMo3-7B-Instruct (\cref{sec:experiments}).

    \item We demonstrate compatibility with an alternative distillation signal by applying RAD to OPD$^2$. 
    This improves four-benchmark mean math accuracy by \textbf{2.66} and \textbf{6.51} percentage points for the 1.7B and 4B Qwen students, respectively (\cref{sec:distillation-compatibility}).
\end{itemize}

\end{mainbox}

\section{Related Work}

\noindent\textbf{On-policy distillation for LLMs.}\enspace
On-policy distillation queries a teacher on student-generated responses, reducing the training--inference mismatch of offline distillation~\citep{agarwal2024policy,gu2024minillm}. Related self-distillation methods obtain teacher signals from the student itself using privileged information such as verified solutions or environment feedback~\citep{shenfeld2026self,hubotter2026reinforcement,zhao2026self}.\\
\noindent\textbf{Unifying distillation and RLVR.}\enspace
KDRL~\citep{xu2025kdrl} \& RLAD~\citep{zhang2026reinforcement} combine verifiable rewards with teacher guidance, while RG-OPD~\citep{rgopd} uses verifier outcomes and teacher likelihoods to gate trajectory-level distillation. ExOPD~\citep{yang2026learning} scales implicit teacher-derived rewards relative to KL regularization.\looseness=-1 \\
\noindent\textbf{Token-selective distillation.}\enspace
TIP~\citep{xu2026tip} selects tokens using uncertainty and teacher--student divergence, while DEAR~\citep{xiao2026finding} identifies decision tokens and supporting evidence. \textsc{Reopold}~\citep{ko2026scalingreasoningefficientlyrelaxed} combines teacher-derived reward clipping with entropy-guided token sampling. 
\citet{armandpour2026unmasking} develop an offline diagnostic of
reward--distillation alignment, estimating token-level $Q$-values 
by repeatedly sampling continuations after selected next-token choices,
which is computationally expensive. They measure
cosine similarity between local logit gradients and study oracle
filtering of positively aligned tokens.
RAD formulates token weighting as a bilevel optimization problem.
A one-step approximation yields the sensitivity of post-update reward
to each token's weight, expressed as parameter-space alignment between
its distillation gradient and an optimizer-adjusted reward direction.
We estimate the reward gradient from minibatch rollouts, avoiding
explicit branchwise $Q$-value estimation. Both the sign and magnitude
of the resulting scores guide projected ascent on bounded, signed
weights, which strengthen, suppress, or reverse distillation
contributions during training.
We compare against representative baselines from both reward-aware and token-selective distillation literature. 
Extended related work appears in Appendix~\ref{app:extended-related-work}.\looseness=-1

\section{Background}
\label{sec:background}

Let \(\bx\sim\mathcal D\) denote a prompt sampled from the data distribution \(\mathcal D\). 
We condition on \(\bx\) and omit its
outer expectation throughout for simplicity. Let \(\pi_{\btheta}\) denote the student model,
\(\by=(y_1,\ldots,y_T)\sim\pi_{\btheta}(\cdot\mid\bx)\) be a response sampled
from the student, and \(s_t=(\bx,y_{<t})\) be the autoregressive state.

\subsection{Reinforcement Learning with Verifiable Rewards}

Reinforcement learning with verifiable rewards (RLVR) trains a policy to
maximize rewards from verifiable outcomes, such as correct answers or passing
code tests. Given a reward function \(R(\bx,\by)\), the expected-reward objective is defined as:
\begin{equation*}
    J_R(\btheta)
    =
    \mathbb E_{\by\sim\pi_{\btheta}(\cdot\mid\bx)}
    \left[R(\bx,\by)\right].
\end{equation*}
Several policy-gradient surrogates can be used to optimize this objective in
RLVR, including proximal policy optimization (PPO)
\citep{schulman2017proximal} and group relative policy optimization (GRPO)
\citep{shao2024deepseekmath}. In this work, we use GRPO's group-relative advantages
to estimate the reward gradient. For each prompt, we sample \(G\) responses
\(\{\by_i\}_{i=1}^{G}\) from the student and normalize their rewards within
the group:
$A_i^R
    =
    \frac{R(\bx,\by_i)-\mu_R}{\sigma_R},
    \mu_R=\frac{1}{G}\sum_{j=1}^{G}R(\bx,\by_j),$
where \(\sigma_R\) is the group reward standard deviation. Every token in \(\by_i\) receives
the same response-level advantage \(A_i^R\).

\subsection{On-Policy Distillation}

On-policy distillation (OPD) samples responses from the student and evaluates
the teacher on the same student-generated states
\citep{agarwal2024policy, gu2024minillm}. For a fixed teacher
\(\pi_{\mathrm{T}}\), the reverse-KL objective is defined as:
\begin{equation*}
    L_D(\btheta)
    =
    \mathbb E_{\by\sim\pi_{\btheta}(\cdot\mid\bx)}
    \left[
        \sum_{t=1}^{T}
        \log
        \frac{\pi_{\btheta}(y_t\mid s_t)}
             {\pi_{\mathrm{T}}(y_t\mid s_t)}
    \right].
\end{equation*}
Unlike GRPO's sequence-level reward, OPD provides dense supervision at every
generated token. Standard OPD, however, weights all token-level teacher
signals equally. Moreover, minimizing \(L_D\) alone provides no monotonic
improvement guarantee for \(J_R\).

\section{Reward-Aligned Distillation}
\label{sec:method}

\methodoverviewfigure

Reward-Aligned Distillation (RAD) selects token-level distillation weights according to their predicted effect on task reward. We introduce the weighted objective (\S\ref{sec:token-weighted-distillation}), then develop a bilevel formulation and its one-step approximation (\S\ref{sec:bilevel}). We extend the resulting scores to general optimizers and compute them efficiently using Jacobian--vector products (\S\ref{sec:general_rad}). Algorithm~\ref{alg:rad} summarizes our training recipe. Finally, we present RAD-Simple, a simplified variant using pre-update reward gradients and shared distillation rollouts (\S\ref{sec:rad-lite}).\looseness=-1

\subsection{Token-Weighted Distillation}
\label{sec:token-weighted-distillation}

Let \(\pi_{\mathrm{old}}\) denote the sampling policy. Holding sampled prefixes fixed, we define the per-token distillation surrogate
\begin{equation*}
    \ell_{D,t}(\btheta;\by)
    =
    \frac{\pi_{\btheta}(y_t\mid s_t)}
         {\pi_{\mathrm{old}}(y_t\mid s_t)}
    \log\frac{\pi_{\btheta}(y_t\mid s_t)}
              {\pi_{\mathrm T}(y_t\mid s_t)}.
\end{equation*}
Assigning real-valued token weights \(\bw(\by)=(w_1(\by),\ldots,w_T(\by))\), with prompt dependence suppressed, gives the weighted objective
\begin{equation}
    L_D(\btheta,\bw)
    =
    \mathbb E_{\by\sim\pi_{\mathrm{old}}(\cdot\mid\bx)}
    \left[
        \sum_{t=1}^{T}w_t(\by)\ell_{D,t}(\btheta;\by)
    \right].
    \label{eq:weighted-rkl}
\end{equation}
The weights can strengthen, suppress, or reverse individual distillation contributions. Uniform weights \(w_t=1\) recover the fixed-prefix OPD surrogate. Appendix~\ref{app:weighted-gradient} derives the corresponding gradient.\looseness=-1

\subsection{Bilevel Token Weighting}\label{sec:bilevel}
Recall that our goal is to choose token weights in order to maximize the reward after distillation. This naturally leads 
to a bilevel optimization formulation:
\begin{equation}
    \max_{\bw}\;J_R\!\left(\btheta^\star(\bw)\right)
    \quad\text{s.t.}\quad
    \btheta^\star(\bw)\in\argmin_{\btheta}L_D(\btheta,\bw).
    \label{eq:bilevel}
\end{equation}
The inner problem trains the student with weighted distillation, and the
outer problem chooses weights to maximize the trained student's reward.
Solving the inner problem for each weight update is computationally
infeasible. Therefore, we approximate its solution with one SGD \citep{robbins1951stochastic} update from the current
student, with learning rate \(\eta>0\):
\begin{equation}
    \btheta^\star(\bw)\approx\btheta^+(\bw)
    =\btheta-\eta\nabla_{\btheta}L_D(\btheta,\bw).
    \label{eq:lookahead}
\end{equation}
At each iteration, we set the initial token weights \(w_t^{(0)}=c\) for every token and 
denote the temporary student
\(\btheta^+=\btheta^+(c\mathbf 1)\).
For the sampled objective, differentiating the post-update reward through
\cref{eq:lookahead} with respect to each token's weight gives the token score,
omitting the loss normalization:
\begin{equation}
    \left.\frac{\partial J_R(\btheta^+(\bw))}{\partial w_t(\by)}
    \right|_{\bw=c\mathbf 1}
    =-\eta\left\langle
        \nabla_{\btheta^+}J_R(\btheta^+),
        \nabla_{\btheta}\ell_{D,t}(\btheta;\by)
    \right\rangle
    \eqqcolon I_t(\by).
    \label{eq:token-score}
\end{equation}
A positive token score favors increasing the token's weight, while a negative score
favors decreasing it. We update the weights by one projected ascent step:
\begin{equation}
    w_t(\by)
    =\Pi_{[c-1,c+1]}\!\left(c+\gamma I_t(\by)\right),
    \qquad \gamma>0.
    \label{eq:projected-weight-update}
\end{equation}
\(\Pi\) clips to the weights to be inside \([c-1,c+1]\) to prevent extreme outlier values 
and \(\gamma\) is the weight learning rate.
We will discuss the choice of initial weight \(c\) in the next section.
We reset the weights to \(c\) for
each new iteration, discard the temporary update, and train the original student
with the updated weights. \Cref{fig:rad-overview} summarizes the iteration.

\begin{algorithm}[t]
\caption{Reward-aligned distillation (RAD)}
\label{alg:rad}
\begin{algorithmic}[1]
\Require Student \(\pi_{\btheta}\), teacher \(\pi_{\mathrm T}\), reward \(R\),
distillation coefficient \(\lambda\), optimizer step \(\mathcal A\),
initial token weight \(c\), weight learning rate \(\gamma\).
\For{each training iteration}
    \State Sample \(\mathcal B=\{(\bx_i,\by_i)\}_i\) from \(\pi_{\btheta}\) and score its tokens with the teacher.
    \State Define weighted distillation loss \(\widehat L_D(\btheta,\bw;\mathcal B)
        =\frac{\lambda}{N}\sum_{i,t}w_{i,t}\ell_{D,t}(\btheta;\by_i)\), where \(N\) is the number of valid response tokens in \(\mathcal B\).
    \State Initialize \(\bw^{(0)}\gets c\mathbf 1\); save parameters and optimizer state.
    \State Take the temporary step
       \(\btheta^+\gets\mathcal A(\btheta,
       \nabla_{\btheta}\widehat L_D(\btheta,c\mathbf 1;\mathcal B))\).
    \State Sample a separate meta batch from the training dataset, generate meta rollouts with
       \(\pi_{\btheta^+}\), and estimate \(\nabla_{\btheta^+}J_R(\btheta^+)\).
    \State Compute \(\boldsymbol p\) using \cref{eq:optimizer-reward-direction}.
    \State Restore the original parameters and optimizer state.
    \State Compute \(\bI\gets-\operatorname{JVP}_{\btheta}
       [\boldsymbol\ell_D(\btheta;\mathcal B)][\boldsymbol p]\).
    \State Update \(w_{i,t}\gets\Pi_{[c-1,c+1]}(c+\gamma I_{i,t})\).
    \State Train the student \(\btheta\gets\mathcal A(\btheta,
       \nabla_{\btheta}\widehat L_D(\btheta,\bw;\mathcal B))\), with detached \(\bw\).
\EndFor
\end{algorithmic}
\end{algorithm}

\subsection{Optimizer-Aware Scoring via Jacobian--Vector Product (JVP)}
\label{sec:general_rad}
We now extend the SGD derivation to general optimizers, including AdamW~\citep{loshchilov2018decoupled}, 
which is widely used in LLM training. This extension accounts for how the student optimizer transforms 
the weighted distillation gradient. 
Let \(\mathcal A(\btheta,\boldsymbol g)\) denote the parameters returned
by one optimizer step with gradient \(\boldsymbol g\), treating the
pre-step optimizer state as fixed. Define
\(\boldsymbol g(\bw)=\nabla_{\btheta}L_D(\btheta,\bw)\).
The one-step approximation becomes
\begin{equation*}
    \max_{\bw}\;J_R\!\left(\btheta^+(\bw)\right)
    \quad\text{s.t.}\quad
    \btheta^+(\bw)=\mathcal A(\btheta,\boldsymbol g(\bw)).
\end{equation*}
We evaluate the token-weight derivatives at \(\bw=c\mathbf1\),
writing \(\boldsymbol g=\boldsymbol g(c\mathbf1)\) and
\(\btheta^+=\btheta^+(c\mathbf1)\).
Similar to \cref{eq:token-score}, differentiating the post-update
reward through this optimizer step gives the
\textit{optimizer-aware token score},
\begin{equation}
    I_t(\by)
    =-\left\langle
        \nabla_{\btheta}\ell_{D,t}(\btheta;\by),
        \boldsymbol p
    \right\rangle,
    \qquad
    \boldsymbol p
    =-\left(
        \frac{\partial\mathcal A(\btheta,\boldsymbol g)}
             {\partial\boldsymbol g}
    \right)^{\!\top}
    \nabla_{\btheta^+}J_R(\btheta^+).
    \label{eq:optimizer-reward-direction}
\end{equation}
For SGD, \(\boldsymbol p=\eta\nabla_{\btheta^+}J_R(\btheta^+)\), recovering
\cref{eq:token-score}. At \(c=0\), the distillation gradient vanishes, so
plain SGD without momentum or weight decay gives \(\btheta^+=\btheta\).
AdamW can still move through its existing moments and decoupled weight decay;
token-weight derivatives can remain nonzero in either case.
For AdamW, we differentiate through the moment updates
and gradient clipping; the derivation is in Appendix~\ref{app:optimizer-hypergradient}.
Let \(\boldsymbol\ell_D\) collect the per-token surrogates. We compute all
scores together using one JVP:
\begin{equation}
    \bI(\by)
    =-\operatorname{JVP}_{\btheta}
    [\boldsymbol\ell_D(\btheta;\by)]
    [\boldsymbol p] \in \R^{\abs{\by}}.
    \label{eq:token-score-jvp}
\end{equation}
The AdamW-specialized JVP is given in \cref{eq:adamw-token-jvp}. We implement
a custom Triton kernel for the flash attention JVP; Appendix~\ref{app:attention-jvp}
gives the derivation and pseudocode (\cref{alg:attention-jvp}).

\noindent\textbf{Choice of initial token weight.}\enspace
The initial token weight \(c\) determines both the temporary distillation update and the starting point for projected ascent. Under the symmetric constraint \([c-1,c+1]\), \(c=1\) starts from uniform weights and learns a weight perturbation around 1 within \([0,2]\), while \(c=0\) starts from zero weights and permits signed weights in \([-1,1]\). We treat \(c\) as a hyperparameter that may vary between iterations. 
Appendix~\ref{app:initial-token-weights} compares these initialization choices. Experiments indicate that \(c=0\) learns slower than \(c=1\) but can achieve better final performance. Gradually expanding the negative range from \(\,[0,1]\,\) to \(\,[-1,1]\,\) for \(c=0\) can help balance learning speed and performance.\looseness=-1

\subsection{RAD Training Algorithm}
\label{sec:batch-implementation}
Distillation rollouts are used to update the student, whereas meta rollouts
are used to estimate the reward gradient for token weighting. Both prompt
batches are sampled from the same training dataset.

RAD's bilevel formulation makes token weighting depend on the student's
post-update reward, introducing additional computation before each weighted
distillation update. At each step, we take a temporary AdamW step with
initial token weights \(c\mathbf 1\), generate separate meta rollouts,
and estimate the reward gradient using GRPO objective. 
We compute the optimizer-aware token scores via a JVP
(\cref{eq:token-score-jvp}), update the weights with
\cref{eq:projected-weight-update}, and train the original student with the
updated weights. The meta batch is only used to learn the token weights; the committed
student update uses the distillation rollouts. Algorithm~\ref{alg:rad}
summarizes the procedure of RAD, and Appendix~\ref{app:implementation-details}
gives the batch objective and implementation details.
RAD can in principle combine any differentiable token-level OPD objective
with any RLVR gradient estimator. \Cref{sec:distillation-compatibility}
demonstrates compatibility with the OPD$^2$~\citep{heo2026policy} distillation signal.

\subsection{RAD-Simple: A Simplified Variant of RAD}
\label{sec:rad-lite}

The bilevel formulation adds a temporary
optimizer update, meta rollouts, a reward-gradient computation, and a JVP
to standard OPD. To simplify token scoring,
\emph{RAD-Simple} estimates the pre-update reward gradient
by reusing the distillation rollouts.
Replacing the post-update reward gradient in \cref{eq:token-score} by
the pre-update reward gradient gives
\begin{equation*}
    I_t(\by)
    =- \left\langle
        \nabla_{\btheta}J_R(\btheta),
        \nabla_{\btheta}\ell_{D,t}(\btheta;\by)
    \right\rangle.
\end{equation*}
We omit the common factor \(\eta\), which can be absorbed into
the weight learning rate \(\gamma\), and use the same projected
weight update in \cref{eq:projected-weight-update}.
RAD-Simple also omits the
optimizer Jacobian, while the student update still uses AdamW.
Reusing the distillation rollouts avoids the temporary student update and
separate meta rollout generation, halving the number of training responses per step.
The only additional computation compared with OPD are
reward-gradient estimation and a JVP. RAD-Simple
ignores changes in the reward gradient after distillation and the
optimizer's effect on token scores. \Cref{tab:main-results} compares
RAD-Simple with RAD at a matched training-response budget.

\begin{table}[ht]
\caption{Math accuracy (\%): RAD at step 100 versus baselines and RAD-Simple at step 200.
Values are mean $\pm$ standard deviation over
16 sampled responses; Mean equally weights six benchmarks.
\textbf{Bold}/\underline{underline} mark the best/second-best trained means, with ties sharing a rank.
Parentheses give gains of the higher RAD or RAD-Simple score in each column
over the best baselines. Time denotes total training time in hours.
}
\label{tab:main-results}
\centering
\small
\renewcommand{\bestmeanstd}[2]{\ensuremath{\mathbf{#1}_{\pm #2}}}
\setlength{\tabcolsep}{0.5pt}
\renewcommand{\arraystretch}{1.0}
\begin{tabularx}{\linewidth}{@{\hspace{\tabcolsep}}l*{7}{>{\centering\arraybackslash}X}>{\centering\arraybackslash}p{3.2em}@{\hspace{\tabcolsep}}}
\toprule
\radhead \multicolumn{1}{c}{Method} & AIME24 & AIME25 & AIME26 & \begin{tabular}[c]{@{}c@{}}HMMT25\\Feb\end{tabular} & \begin{tabular}[c]{@{}c@{}}HMMT25\\Nov\end{tabular} & \begin{tabular}[c]{@{}c@{}}Olympiad\\Bench\end{tabular} & Mean & \shortstack{Time\\(h)} \\
\midrule
Teacher & \meanstd{74.79}{3.72} & \meanstd{61.46}{3.53} & \meanstd{74.79}{4.56} & \meanstd{43.75}{5.38} & \meanstd{59.17}{5.59} & \meanstd{77.15}{0.52} & \meanstd{65.18}{1.66} & --- \\
\midrule
\radhead\multicolumn{9}{l}{\textit{\textbf{Student: Qwen3-1.7B (non-thinking)}}} \\
\addlinespace[1pt]
Base & \meanstd{12.08}{3.89} & \meanstd{11.04}{4.20} & \meanstd{8.13}{2.63} & \meanstd{6.04}{3.38} & \meanstd{3.75}{2.00} & \meanstd{39.30}{0.82} & \meanstd{13.39}{1.49} & --- \\
\addlinespace[1pt]
GRPO & \meanstd{29.58}{5.64} & \meanstd{19.58}{3.89} & \meanstd{18.96}{5.98} & \meanstd{15.63}{2.82} & \meanstd{10.21}{4.93} & \meanstd{53.07}{0.93} & \meanstd{24.51}{1.46} & 5.12 \\
OPD & \meanstd{40.00}{5.65} & \meanstd{27.29}{4.44} & \meanstd{26.46}{5.83} & \meanstd{17.71}{3.67} & \meanstd{16.67}{6.24} & \meanstd{58.19}{1.17} & \meanstd{31.05}{2.01} & 14.28 \\
TIP & \meanstd{39.17}{4.93} & \meanstd{28.54}{4.40} & \meanstd{31.04}{5.98} & \meanstd{17.08}{3.09} & \meanstd{18.33}{4.08} & \meanstd{57.93}{1.18} & \meanstd{32.02}{1.70} & 34.09 \\
RLAD & \meanstd{35.21}{4.08} & \meanstd{27.71}{4.20} & \meanstd{26.04}{6.89} & \meanstd{16.88}{3.43} & \meanstd{17.08}{4.55} & \meanstd{58.06}{1.27} & \meanstd{30.16}{1.58} & 14.11 \\
ExOPD & \meanstd{36.88}{4.33} & \meanstd{31.67}{4.08} & \meanstd{31.25}{5.51} & \meanstd{16.67}{2.89} & \meanstd{19.38}{5.43} & \meanstd{59.69}{1.12} & \meanstd{32.59}{1.81} & 51.02 \\
\midrule
\radrow\radlabel{\method{}} & \bestmeanstd{47.50}{7.41} & \bestmeanstd{36.04}{3.58} & \bestmeanstd{38.96}{3.86} & \bestmeanstd{22.50}{4.17} & \bestmeanstd{24.79}{4.08} & \bestmeanstd{64.60}{1.05} & \bestmeanstd{39.07}{1.59} & 16.66 \\
\radrow\radlabel{RAD-Simple} & \secondmeanstd{46.88}{5.83} & \secondmeanstd{34.58}{5.64} & \secondmeanstd{38.54}{5.27} & \secondmeanstd{21.67}{3.54} & \secondmeanstd{24.38}{3.05} & \secondmeanstd{64.47}{1.07} & \secondmeanstd{38.42}{1.70} & 32.48 \\
\radgainrow \parbox[c]{6em}{\raggedright\baselinegainlabel} & \radgain{+7.50} & \radgain{+4.38} & \radgain{+7.71} & \radgain{+4.79} & \radgain{+5.42} & \radgain{+4.92} & \radgain{+6.48} & --- \\
\midrule
\radhead\multicolumn{9}{l}{\textit{\textbf{Student: Qwen3-4B-Instruct-2507}}} \\
\addlinespace[1pt]
Base & \meanstd{64.17}{5.95} & \meanstd{48.33}{5.53} & \meanstd{53.13}{5.46} & \meanstd{32.29}{4.52} & \meanstd{44.58}{4.06} & \meanstd{71.77}{0.96} & \meanstd{52.38}{1.47} & --- \\
\addlinespace[1pt]
GRPO & \meanstd{67.92}{6.44} & \meanstd{55.63}{3.67} & \secondmeanstd{62.29}{4.52} & \meanstd{34.58}{3.51} & \meanstd{48.75}{5.38} & \meanstd{74.58}{0.84} & \meanstd{57.29}{1.72} & 14.77 \\
OPD & \meanstd{65.21}{6.23} & \meanstd{50.63}{4.60} & \meanstd{55.42}{5.25} & \meanstd{33.96}{3.95} & \meanstd{41.46}{5.00} & \meanstd{71.89}{0.74} & \meanstd{53.09}{1.72} & 12.47 \\
TIP & \meanstd{61.04}{4.82} & \meanstd{49.79}{5.71} & \meanstd{53.96}{6.79} & \meanstd{33.96}{4.44} & \meanstd{41.04}{6.09} & \meanstd{71.77}{0.83} & \meanstd{51.93}{1.85} & 24.46 \\
RLAD & \meanstd{59.79}{5.06} & \meanstd{47.71}{4.96} & \meanstd{53.33}{6.12} & \meanstd{32.29}{4.67} & \meanstd{39.58}{5.64} & \meanstd{72.50}{0.93} & \meanstd{50.87}{1.60} & 13.30 \\
ExOPD & \meanstd{58.13}{5.13} & \meanstd{50.42}{3.51} & \meanstd{55.21}{5.52} & \meanstd{33.33}{4.08} & \meanstd{40.63}{5.17} & \meanstd{71.26}{0.92} & \meanstd{51.49}{2.02} & 39.61 \\
\midrule
\radrow\radlabel{\method{}} & \secondmeanstd{71.25}{5.51} & \bestmeanstd{62.50}{3.63} & \secondmeanstd{62.29}{5.23} & \secondmeanstd{36.04}{5.30} & \secondmeanstd{51.25}{5.12} & \secondmeanstd{75.46}{0.87} & \secondmeanstd{59.80}{2.07} & 26.79 \\
\radrow\radlabel{RAD-Simple} & \bestmeanstd{71.88}{5.13} & \secondmeanstd{60.63}{3.77} & \bestmeanstd{63.96}{5.17} & \bestmeanstd{40.00}{4.41} & \bestmeanstd{52.29}{5.36} & \bestmeanstd{76.30}{0.79} & \bestmeanstd{60.84}{1.50} & 27.88 \\
\radgainrow \parbox[c]{6em}{\raggedright\baselinegainlabel} & \radgain{+3.96} & \radgain{+6.88} & \radgain{+1.67} & \radgain{+5.42} & \radgain{+3.54} & \radgain{+1.71} & \radgain{+3.55} & --- \\
\bottomrule
\end{tabularx}
\end{table}

\section{Experiments}
\label{sec:experiments}

We describe the experimental setup and evaluation protocol, then present math and coding results. Implementation details, including inference hyperparameters, appear in Appendix~\ref{app:experimental-details}.

\subsection{Experimental Settings}
\label{sec:experimental-settings}

We evaluate RAD on mathematical reasoning and code generation, and additionally evaluate RAD-Simple on mathematical reasoning.

\noindent\textbf{Models.}\enspace
For mathematical reasoning, we use Qwen3-30B-A3B-Instruct-2507 as the teacher and Qwen3-1.7B (non-thinking) and Qwen3-4B-Instruct-2507 as students~\citep{yang2025qwen3}. For code generation, we evaluate Qwen3-1.7B (non-thinking) with the same teacher, and OLMo3-7B-Instruct with OLMo3.1-32B-Instruct as its teacher.

\noindent\textbf{Training protocol.}\enspace
We train with \texttt{verl}~\citep{sheng2025hybridflow} using 64 prompts and eight responses per prompt. Within each domain and student setting, all methods share the training data, student initialization, rollout group size, sampling configuration, and optimizer settings. RAD additionally samples an equally sized, separate meta batch from the temporary student and updates token weights using exact JVPs. We compare 100 RAD updates with 200 baseline updates, matching the generated training responses, including RAD's meta rollouts. At this budget, RAD uses 17--18\% more training time than OPD and RLAD on Qwen3-1.7B math, but less than TIP and ExOPD. On OLMo3-7B coding, RAD uses 18--39\% less training time than the reported baselines (\cref{tab:training-step-comparison}). RAD uses the initial token weight \(c=0\) in the main math and coding comparisons, with an additional cosine schedule of negative range for coding. Shared and method-specific settings appear in Appendix~\ref{app:training-settings}.

\noindent\textbf{Baselines.}\enspace
We compare RAD and RAD-Simple with GRPO~\citep{shao2024deepseekmath}, OPD~\citep{agarwal2024policy}, TIP~\citep{xu2026tip}, RLAD~\citep{zhang2026reinforcement}, and ExOPD~\citep{yang2026learning}. Given the higher training cost of the OLMo3-7B setting (\cref{tab:training-step-comparison}), we retain GRPO and OPD as reference methods, with TIP and RLAD representing strongest token-selective and reward-aware distillation baselines, respectively. Baseline-specific hyperparameters follow the respective papers.

\noindent\textbf{Mathematical reasoning.}\enspace
Following ExOPD~\citep{yang2026learning}, we train on 57,046 problems from DeepMath-103K~\citep{he2025deepmath} with difficulty ratings of at least 6. We evaluate on AIME24~\citep{Aime2024}, AIME25~\citep{Aime2025}, AIME26, HMMT2025-Feb, HMMT2025-Nov~\citep{balunovic2026matharena}, and OlympiadBench~\citep{he2024olympiadbench}. We sample 16 responses per problem at temperature 1.0 and top-\(p\) 1.0, with a 32,768-token generation limit. Responses are graded with Math-Verify~\footnote{https://github.com/huggingface/Math-Verify}, and pass@1 is estimated as the fraction of sampled responses that are correct. The overall score is the unweighted mean across the six benchmarks.

\noindent\textbf{Code generation.}\enspace
We train on the Eurus code split~\citep{cui2025process}, assigning reward 1 when all tests pass and 0 otherwise. We evaluate on all 1,055 LiveCodeBench-v6 tasks~\citep{jain2025livecodebench}. Following ExOPD~\citep{yang2026learning}, we sample four responses per task at temperature 1.0 and top-\(p\) 1.0, with a 32,768-token generation limit. Dataset statistics appear in \cref{tab:dataset-statistics}.

\subsection{Mathematical Reasoning Results}

\begin{figure}[tbp]
\centering
\includegraphics[width=0.95\linewidth,trim=0 12bp 0 0,clip]{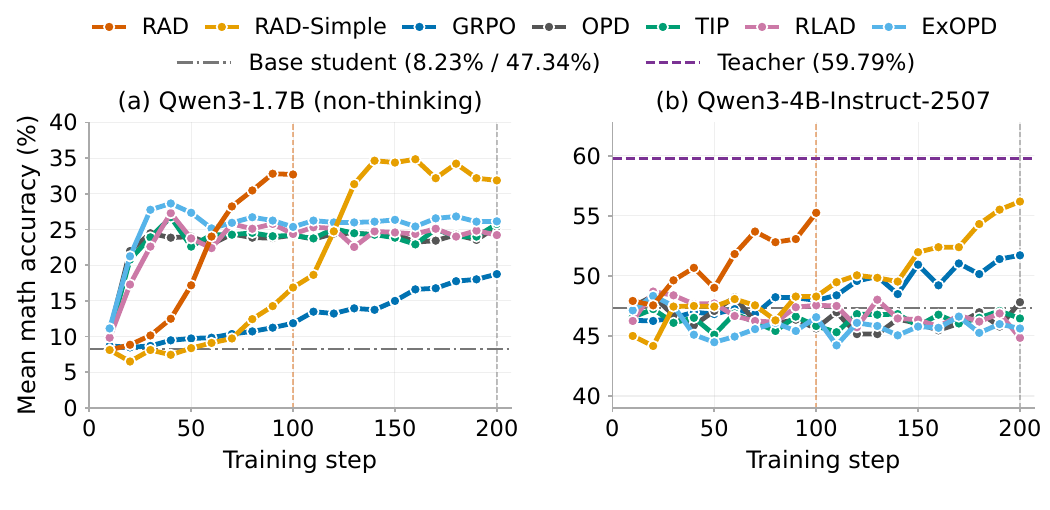}
\captionsetup{skip=5pt}
\looseness=-1
\caption{Math evaluation dynamics for Qwen3-1.7B (non-thinking) in the left panel and Qwen3-4B-Instruct-2507 in the right panel, averaged equally over AIME24/25 and HMMT2025-Feb/Nov. Distillation baselines tend to plateau or decline early, while GRPO improves slowly. RAD reaches a higher peak within 100 updates than these baselines achieve over 200 updates. 
}
\label{fig:math-eval-dynamics}
\end{figure}

At the matched training-response budget (\cref{tab:main-results}), RAD reaches \textbf{39.07\%} mean accuracy with Qwen3-1.7B (non-thinking), exceeding the strongest
baseline, ExOPD at 32.59\%, by \textbf{6.48} points. With Qwen3-4B-Instruct-2507,
RAD reaches \textbf{59.80\%}, exceeding the strongest baseline, GRPO at 57.29\%,
by \textbf{2.51} points. The corresponding gains over OPD are \textbf{8.01} and \textbf{6.71 points}.
\Cref{fig:math-eval-dynamics} tracks both students on four
benchmarks (AIME24/25 and HMMT2025-Feb/Nov).
For the 1.7B student, RAD improves more slowly
early in training but continues to improve as the distillation
baselines plateau or decline. For the 4B student, RAD exceeds
all baselines from step 30 onward.
RAD-Simple trails RAD at 100 updates for both students but continues to improve,
reaching comparable accuracy at 200 updates.\looseness=-1

At the same training-response budget, RAD-Simple reaches
\(38.42\pm1.70\%\) and \(60.84\pm1.50\%\) mean accuracy after 200 updates
for the 1.7B and 4B students, respectively (\cref{tab:main-results}).
These results are \textbf{0.65} points below and \textbf{1.04} points above RAD at step 100, respectively, and exceed the strongest baselines by
\textbf{5.83} and \textbf{3.55} points. This shows that the simpler
pre-update scores can also support effective token weighting.
However, within the same 100 training steps, RAD-Simple learns slower than RAD as shown in Figure~\ref{fig:math-eval-dynamics}.
RAD-Simple removes the temporary optimizer update and separate meta rollouts, and computes token scores using the current-policy reward gradient without the optimizer Jacobian. It achieves similar math accuracy at a matched generated-response budget, but requires more training time for Qwen3-1.7B. This trade-off supports the practical value of RAD’s full scoring procedure: additional computation for estimating token weights can improve overall training efficiency.\looseness=-1

For Qwen3-4B-Instruct-2507, OPD reaches 53.09\% at step 200,
slightly above the base student's 52.38\%, while TIP, RLAD, and
ExOPD perform below the base student.
Similar degradation of OPD and ExOPD is observed in \cite{heo2026policy} (Table 4 and 5).
RAD's gain in this setting supports using post-update reward to guide the contribution of each teacher signal even when the initial teacher--student gap is small.

Appendix~\ref{app:math-training-time} additionally compares RAD with baselines with a matched training time.
\Cref{fig:rad100-grpo-extended} in Appendix~\ref{app:extended-grpo} compares RAD at step 90 with GRPO at step 450 on math and coding, with GRPO using \textbf{2.5$\times$} as many generated training responses.
Appendix~\ref{app:training-time} compares total training times for Qwen3-1.7B (non-thinking) on math and OLMo3-7B-Instruct on coding, and reports a RAD training-time breakdown for Qwen3-1.7B (non-thinking).\looseness=-1

\subsection{Code Generation Results}

\begin{figure}[ht]
\centering
\includegraphics[width=0.95\linewidth]{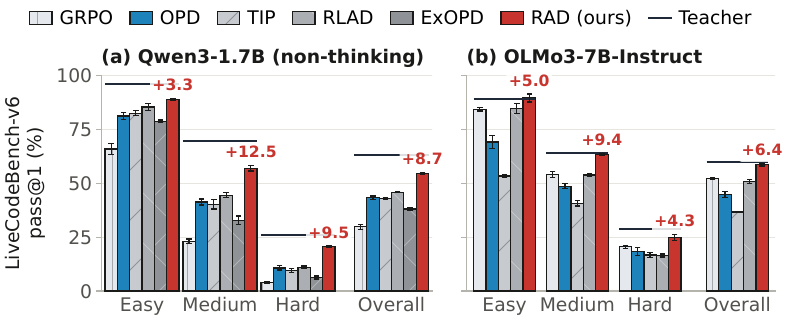}
\caption{LiveCodeBench-v6 pass@1 (\%) by difficulty, comparing RAD at step
100 with baselines at step 200. RAD outperforms the baselines across all difficulty levels for both student models. Numbers above RAD's bars give its
improvement over the best baseline in that tier (percentage points). Error bars show
$\pm$1 standard deviation over four sampled responses; black lines mark available teacher means. 
\Cref{tab:coding-results} reports the full means and standard deviations.}
\label{fig:coding-results-bars}
\end{figure}

\Cref{fig:coding-results-bars} compares both students across varying difficulty tiers on full LiveCodeBench-v6.
Appendix~\ref{app:lcb-results} reports the full numerical results
(\cref{tab:coding-results}). 
RAD with Qwen3-1.7B (non-thinking) reaches
\(\bf 54.69\pm0.31\)\% pass@1 at step 100, compared with
\(45.97\pm0.26\)\% for RLAD, the strongest step-200 baseline. RAD also leads
the five evaluated baselines on Easy, Medium, and Hard.
Appendix~\ref{app:coding-validation}
reports Qwen3-1.7B (non-thinking) coding-validation dynamics and checkpoint
results.

\label{sec:olmo-extension}

With OLMo3-7B-Instruct, RAD reaches \(\bf 58.67\pm0.52\)\% on full
LiveCodeBench-v6 at step 100, exceeding GRPO at \(52.27\pm0.26\)\%,
the strongest completed step-200 baseline, by 6.40 points. These results extend the coding comparison to a
different model family and a larger student.
Appendix~\ref{app:olmo-coding}
reports OLMo3-7B-Instruct coding-validation dynamics and checkpoint
results.

Preliminary experiments also extend RAD to agentic software engineering.
On SWE-bench Verified, RAD achieves 50.5\% pass@1, outperforming the
strongest baseline by 3.7 percentage points while training for half as
many epochs (Appendix~\ref{app:agentic}).

\section{Analysis and Extensions}
\label{sec:analysis-and-extensions}

\begin{table}[t]
\caption{Qwen3-1.7B (non-thinking) math ablations at step 100 ($c=0$): mean
accuracy over the four benchmarks (AIME24/25 and HMMT2025-Feb/Nov).
Reversing scores or shuffling weights sharply reduces accuracy.
Variants use separate meta rollouts unless rollout reuse is specified.
$\Delta$ (pp) denotes difference from RAD.
\textbf{Bold}/\underline{underline}: best/second-best displayed means.
}
\label{tab:math-ablations}
\centering
\normalsize
\setlength{\tabcolsep}{3pt}
\renewcommand{\arraystretch}{1.0}
\begin{tabularx}{\linewidth}{@{}>{\raggedright\arraybackslash}p{0.27\linewidth}>{\raggedright\arraybackslash}Xrr@{}}
\toprule
\radhead Variant & Change from RAD & Acc. (\%) & $\Delta$ (pp) \\
\midrule
\radrow\radlabel{RAD} & AdamW lookahead; separate meta rollouts & \underline{32.71} & --- \\
Negative score & Reverse the score: $I_t\mapsto -I_t$ & 0.36 & $(-32.34)$ \\
Shuffled weights & Permute weights within each response & 6.35 & $(-26.35)$ \\
\midrule
Positive only & Restrict weights to $[0,1]$ & 25.00 & $(-7.71)$ \\
Negative-bound warmup & Cosine lower bound: $0\to -1$ over 50 steps & 31.20 & $(-1.51)$ \\
\midrule
Pre-update gradient & Reward gradient before the temporary update & \textbf{34.32} & $(+1.61)$ \\
 ~~+ rollout reuse & Pre-update reward grad; reuse distillation rollouts & 18.54 & $(-14.17)$ \\
\midrule
SGD lookahead & SGD lookahead update & 16.88 & $(-15.83)$ \\
\bottomrule
\end{tabularx}
\end{table}

In this section, we examine which components drive RAD's performance, how token reweighting
affects gradient alignment, and whether RAD extends to other distillation signals.

\subsection{RQ1: Which components contribute to RAD's performance?}
\label{sec:ablations}

We ablate RAD using Qwen3-1.7B (non-thinking) with \(c=0\) after 100 updates on AIME24/25 and HMMT2025-Feb/Nov (\cref{tab:math-ablations}). Reversing scores or shuffling weights reduces accuracy from 32.71\% to 0.36\% and 6.35\%, respectively, supporting token-specific alignment. Restricting weights to \([0,1]\) yields 25.00\%, while gradually introducing negative weights achieves 31.20\%. Replacing AdamW lookahead with SGD yields 16.88\%. These comparisons support signed weights and optimizer-aware scoring in this setting.

Evaluating the reward gradient before the temporary update achieves 34.32\%, comparable to RAD.
However, it achieves 47.19\% compared with 55.26\% of RAD for Qwen3-4B-Instruct-2507, 
showing the benefit depends on the student setting.
Reusing distillation rollouts further reduces accuracy to 18.54\%, but also halves the generated responses per update. We next examine how token reweighting changes gradient alignment in RAD and RAD-Simple.

\subsection{RQ2: Does token reweighting improve gradient alignment?}
\label{sec:gradient-alignment}

\begin{table}[t]
\centering
\begin{minipage}[t]{0.51\linewidth}
\vspace{0pt}
\centering
\caption{Cosine between distillation descent direction and the reward direction. Uniform uses \(\bw=\mathbf1\); reweighted uses the learned token weights.}
\label{tab:gradient-alignment}
\normalsize
\setlength{\tabcolsep}{3pt}
\begin{tabular*}{\linewidth}{@{\extracolsep{\fill}}lrrr@{}}
\toprule
Method & Step & Uniform & Reweighted \\
\midrule
\multicolumn{4}{@{}l}{\textit{Cosine with \(\boldsymbol p\)}} \\
RAD & 0 & \(-0.1437\) & \(0.4487\) \\
RAD & 100 & \(0.0004\) & \(0.0158\) \\
\midrule
\multicolumn{4}{@{}l}{\textit{Cosine with \(\nabla_{\btheta}J_R(\btheta)\)}} \\
RAD & 0 & \(0.1404\) & \(-0.0658\) \\
RAD & 100 & \(0.0687\) & \(0.0357\) \\
RAD-Simple & 0 & \(0.0771\) & \(0.4816\) \\
RAD-Simple & 100 & \(-0.0008\) & \(0.4508\) \\
\bottomrule
\end{tabular*}
\end{minipage}\hfill
\begin{minipage}[t]{0.47\linewidth}
\vspace{0pt}
\caption{RAD compatibility with OPD$^2$: RAD at step 100 versus baselines
at step 200. Values are mean
math accuracy over AIME24/25 and
HMMT2025-Feb/Nov.}
\label{tab:opd2-compatibility}
\centering
\normalsize
\setlength{\tabcolsep}{1.5pt}
\renewcommand{\arraystretch}{1.05}
\begin{tabular*}{\linewidth}{@{\extracolsep{\fill}}lcc@{}}
\toprule
Method & \shortstack{Qwen3-1.7B\\(non-thinking)} & \shortstack{Qwen3-4B-\\Instruct-2507} \\
\midrule
Teacher & 59.79 & 59.79 \\
\midrule
Base & 8.23 & 47.34 \\
\addlinespace[1pt]
Uniform OPD & 25.42 & 47.81 \\
OPD$^2$ & \underline{28.33} & \underline{53.13} \\
OPD$^2$ + RAD & \textbf{30.99} & \textbf{59.64} \\
\midrule
\shortstack[l]{$\Delta$ vs.\ OPD$^2$} & +2.66 & +6.51 \\
\bottomrule
\end{tabular*}

\end{minipage}
\end{table}

We compare uniform weights \(\bw=\mathbf1\) and learned token weights at fixed Qwen3-1.7B
(non-thinking) math checkpoints at steps 0 and 100. We measure cosine alignment
between distillation descent direction (negative distillation gradient), \(-\nabla_{\btheta}L_D(\btheta,\bw)\), and 
RAD's optimizer-aware reward direction
\(\boldsymbol p\) (\cref{eq:optimizer-reward-direction}), and current
student's reward direction \(\nabla_{\btheta}J_R(\btheta)\).

\Cref{tab:gradient-alignment} shows that reweighting increases cosine with 
\(\boldsymbol p\) for RAD and \(\nabla_{\btheta}J_R(\btheta)\) for RAD-Simple
at both checkpoints. RAD's cosine with
\(\nabla_{\btheta}J_R(\btheta)\) decreases: 
RAD’s token weights target the optimizer-aware reward direction \(\boldsymbol p\), which accounts for how changes in the distillation gradient affect post-update reward, rather than the immediate reward gradient.

\subsection{RQ3: Is RAD compatible with other distillation signals?}
\label{sec:distillation-compatibility}

\begin{wrapfigure}{r}{0.50\linewidth}
\centering
\vspace{-\intextsep}
\includegraphics[width=\linewidth]{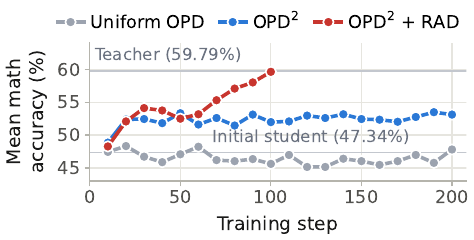}
\captionsetup{skip=4pt,justification=raggedright,singlelinecheck=false}
\caption{OPD$^2$ evaluation dynamics for Qwen3-4B-Instruct-2507, using
\cref{tab:opd2-compatibility}'s benchmarks. Horizontal lines mark the initial student and teacher. 
OPD$^2$ + RAD approaches teacher accuracy at step 100.
}
\label{fig:opd2-4b-eval}
\end{wrapfigure}

We apply RAD's token weighting to OPD$^2$~\citep{heo2026policy}, which derives supervision 
from an instruction-tuned teacher and its paired base model. 
We evaluate on the same four math benchmarks; implementation details appear in Appendix~\ref{app:training-settings}.

\Cref{tab:opd2-compatibility} compares OPD$^2$ + RAD after 100 updates with uniform OPD and OPD$^2$ after 200 updates, 
matching the number of generated training responses. Adding RAD improves OPD$^2$, supporting its compatibility with 
an alternative distillation signal. For Qwen3-4B-Instruct-2507, the combined method also approaches 
teacher accuracy (\cref{fig:opd2-4b-eval}).

\clearpage

\section{Conclusion and Future Work}

\method{} uses optimizer-aware lookahead scores and projected ascent to
align token-level distillation with reward. It improves math and coding
accuracy at a matched training-response budget. Results with
OLMo3-7B-Instruct extend the evidence to a different model family and a
larger student, while OPD$^2$ results support compatibility with other
distillation signals.
RAD-Simple reuses distillation rollouts and removes the temporary update;
it improves over the strongest math baselines for both Qwen students at the same
training-response budget, demonstrating the effectiveness of simpler
token scoring in these settings.

\noindent\textbf{Limitations and future work.}\enspace
RAD requires a temporary student update, fresh outer rollouts, and an exact JVP, increasing its per-step cost relative to standard OPD. RAD-Simple removes the temporary update and fresh outer rollouts but still requires reward-gradient estimation and a JVP. Further reducing these costs and understanding when optimizer-aware lookahead is beneficial remain important directions.
Future work should explore longer training horizons, and learned or non-verifiable rewards.

\bibliography{ref.bib}
\bibliographystyle{iclr2027_conference}

\newpage
\appendix
\onecolumn
\appendixpage

\begingroup
\setcounter{tocdepth}{2}
\tableofcontents
\endgroup
\clearpage

\section{Experimental Details and Additional Math Results}
\label{app:experimental-details}

\subsection{Training Settings}
\label{app:training-settings}

\noindent\textbf{Shared training setup.}\enspace
The reported main math and coding experiments use eight NVIDIA H100 GPUs.
Each update uses 64 actor prompts with eight responses per prompt; RAD
additionally samples 64 separate meta prompts with eight responses each.
\Cref{tab:shared-training-hparams,tab:method-hparams} summarize the shared
and method-specific math hyperparameters. We describe the datasets first,
followed by the baseline implementations and RAD weight settings.

\noindent\textbf{Math datasets and evaluation.}\enspace
Dataset sizes and evaluation metrics are listed in
\cref{tab:dataset-statistics}. For AIME26, we use the 30-problem
\href{https://huggingface.co/datasets/MathArena/aime_2026}{MathArena snapshot}
with two \href{https://github.com/eth-sri/matharena/issues/31}{upstream-confirmed transcription corrections}. For OlympiadBench,
we use the 675-problem \href{https://github.com/QwenLM/Qwen2.5-Math/blob/a45202bd16f1ec06f433442dc1152d0074773465/evaluation/data/olympiadbench/test.jsonl}{Qwen2.5-Math evaluation snapshot}.

\noindent\textbf{Coding data.}\enspace
The Eurus code split is assembled from APPS \citep{hendrycks2021apps},
CodeContests \citep{li2022competition}, Codeforces \citep{codeforces},
and TACO \citep{li2023taco}. The coding results in
\cref{tab:coding-results,tab:coding-validation-results,tab:olmo-coding-results}
use 25,276 training examples and 1,024 validation examples
(\cref{tab:dataset-statistics}).

\noindent\textbf{Coding evaluation.}\enspace
LiveCodeBench-v6 contains 322 easy, 383 medium, and 350 hard tasks.
Evaluation uses a 40,960-token context limit for the combined prompt and
response, with at most 32,768 generated tokens per response. The main
coding comparison reports RAD at step 100 and baselines at step 200.
\begin{table}[htbp]
\caption{Dataset statistics and evaluation metrics. Counts denote dataset
examples or tasks; responses per prompt are sampled whenever a prompt is
selected. RAD additionally samples an independent outer batch of 64 meta
prompts with eight responses each (512 responses). Coding training and
validation counts refer to the Eurus split before model-specific
prompt-length filtering.}
\label{tab:dataset-statistics}
\centering
\normalsize
\setlength{\tabcolsep}{3pt}
\renewcommand{\arraystretch}{1.0}
\begin{tabularx}{\linewidth}{@{}ll>{\raggedright\arraybackslash}XrC{0.12\linewidth}>{\raggedright\arraybackslash}X@{}}
\toprule
Domain & Use & Dataset & \# Problems & Responses / prompt & Metric \\
\midrule
Math & Train & DeepMath-103K, level \(\geq 6\)
& 57,046 & 8 & Binary correctness reward \\
Math & Eval & AIME24 & 30 & 16
& pass@1 accuracy (Math-Verify) \\
Math & Eval & AIME25 & 30 & 16
& pass@1 accuracy (Math-Verify) \\
Math & Eval & AIME26 & 30 & 16
& pass@1 accuracy (Math-Verify) \\
Math & Eval & HMMT2025-Feb & 30 & 16
& pass@1 accuracy (Math-Verify) \\
Math & Eval & HMMT2025-Nov & 30 & 16
& pass@1 accuracy (Math-Verify) \\
Math & Eval & OlympiadBench & 675 & 16
& pass@1 accuracy (Math-Verify) \\
\midrule
Code & Train & Eurus code split (APPS, CodeContests, Codeforces, and TACO)
& 25,276 & 8 & Binary all-tests-pass reward \\
Code & Val & APPS & 142 & 1 & pass@1 functional correctness \\
Code & Val & CodeContests & 377 & 1 & pass@1 functional correctness \\
Code & Val & Codeforces & 123 & 1 & pass@1 functional correctness \\
Code & Val & TACO & 382 & 1 & pass@1 functional correctness \\
Code & Eval & LiveCodeBench-v6
& 1,055 & 4 & pass@1 functional correctness (native execution) \\
\bottomrule
\end{tabularx}
\end{table}

\noindent\textbf{Baseline implementations.}\enspace
GRPO optimizes task reward without teacher guidance, while OPD applies
reverse-KL distillation uniformly to all response tokens. TIP uses Soft-OR
selection with 50\% token retention. RLAD modifies the clipped GRPO update
with a teacher-aware policy ratio. ExOPD constructs a reward-extrapolated
reverse-KL advantage using frozen student and teacher base policies.
OPD, RLAD, ExOPD, and RAD use sampled-action estimates of reverse KL,
while TIP uses full-vocabulary reverse KL as in their paper.

\noindent\textbf{RAD weight settings.}\enspace
For the math and coding experiments, RAD uses distillation coefficient
\(\lambda=0.1\) to prevent excessively frequent gradient clipping, and weight learning rate \(\gamma=10^4 N_1/N\)
in \cref{eq:projected-weight-update}. Here \(N_1\) and \(N\) are the
numbers of valid response tokens in the first and current actor batches,
respectively.
The unprojected weight change is \(\gamma I_t\): the score \(I_t\) already
includes the optimizer learning rate through the optimizer Jacobian and
the scaling of the estimated outer reward gradient. The weight learning rate
should therefore be interpreted together with this score scale;
the resulting weights are clipped to the specified interval.
We apply no further weight normalization. Both math runs use the fixed initial
token weight \(c=0\) and clipping interval \([-1,1]\). These are experimental
settings, while \cref{eq:projected-weight-update} keeps \(c\) configurable.
For RAD-Simple on math, we use \(\lambda=0.1\) and
\(\gamma=3N_1/N\) for both Qwen3-1.7B (non-thinking) and Qwen3-4B-Instruct-2507.
Both use \(c=0\) and clipping interval \([-1,1]\) throughout 200 updates.

The Qwen3-1.7B (non-thinking) coding RAD run uses the same weight learning rate
\(\gamma=10^4 N_1/N\) with distillation coefficient \(\lambda=0.1\),
without further weight normalization. It keeps the teacher's thinking prefix,
uses the initial token weight \(c=0\), and clips weights to an upper bound of 1 and a lower
bound that follows a cosine schedule from 0 to \(-1\) over the first 50 steps.
The initial token weight stays fixed throughout this clipping-bound schedule,
which gradually permits negative weights. The OLMo coding run uses the same
initial token weight and clipping schedule (Appendix~\ref{app:olmo-coding}).

\noindent\textbf{OPD$^2$ compatibility settings.}\enspace
For the OPD$^2$ compatibility experiment, the teacher pair is
Qwen3-30B-A3B-Instruct-2507 and Qwen3-30B-A3B-Base. Our local
implementation uses frozen-student TopK64 support and global token-mean
reduction, with distillation coefficient \(\lambda=0.1\) for both
OPD$^2$ and OPD$^2$ + RAD.
The RAD variant uses a cosine schedule for the initial token weight \(c\)
from 1 to 0, a clipping radius of 1 around \(c\), and weight learning rate
\(\gamma=10^4 N_1/N\). This schedule changes both
the initial token weight and the clipping interval, moving the latter from
\([0,2]\) to \([-1,1]\). Each step uses 64 actor
prompts with eight responses each; RAD additionally uses 64 meta prompts
with eight responses each. We compare OPD$^2$ + RAD at step 100 with
uniform OPD and OPD$^2$ at step 200.

\subsection{Implementation Details}
\label{app:implementation-details}

\noindent\textbf{Batch objective.}\enspace
Let \(\mathcal B=\{(\bx_i,\by_i)\}_i\) be the current student rollout
batch, with prefixes \(s_{i,t}=(\bx_i,y_{i,<t})\), and let \(N\) be its
number of valid response tokens. The temporary and committed updates use
the normalized batch objective
\begin{equation}
    \widehat L_D(\btheta,\bw;\mathcal B)
    =\frac{\lambda}{N}\sum_{i,t}w_{i,t}\ell_{D,t}(\btheta;\by_i),
    \label{eq:batch-rkl}
\end{equation}
where \(\lambda\) is the distillation coefficient. We keep the loss normalization
fixed when differentiating with respect to the weights. The exact weight
derivative is \((\lambda/N)I_{i,t}\)
(\cref{eq:general-optimizer-hypergradient}). We update the weights using
the scores \(I_{i,t}\) directly with the separately specified weight learning
rate \(\gamma\) in \cref{eq:projected-weight-update}.
We also apply a detached token-level
rollout/training importance correction clipped at \(5\), omitted from the
notation. The weights are detached for the committed student update.

\noindent\textbf{JVP backends.}\enspace
The \texttt{flash\_triton} attention implementation combines a FlashAttention
forward pass with a custom Triton kernel for the attention JVP
(\cref{alg:attention-jvp}). The Qwen3-1.7B (non-thinking) and
Qwen3-4B-Instruct-2507 RAD experiments use this implementation to compute
the model JVP on a temporary full student replica with its parameter direction.
The \texttt{olmo3\_layer\_streamed} backend, used for OLMo3-7B-Instruct RAD
coding, computes the model JVP one layer at a time: it gathers each parameter
group and its direction, propagates activations and their tangents, and releases
the gathered tensors after use. This execution strategy reduces parameter and
direction storage while reusing the same \texttt{flash\_triton} attention rule;
Appendix~\ref{app:layer-streamed-jvp} gives the details.

\begin{center}
\begin{minipage}[t]{0.47\linewidth}
\captionof{table}{Shared math training hyperparameters.}
\label{tab:shared-training-hparams}
\centering
\normalsize
\setlength{\tabcolsep}{5pt}
\renewcommand{\arraystretch}{1.0}
\begin{tabular}{@{}C{0.66\linewidth}c@{}}
\toprule
Hyperparameter & Value \\
\midrule
Training prompts per step & 64 \\
Rollouts per prompt & 8 \\
Actor epochs & 1 \\
Maximum prompt length & 2,048 \\
Maximum response length & 32,768 \\
Training temperature & 1.0 \\
Training top-\(p\) & 1.0 \\
Optimizer & AdamW \\
Learning rate & \(1\times10^{-6}\) \\
Weight decay & 0.01 \\
Global gradient clipping & 1.0 \\
Random seed & 42 \\
RAD / baseline steps & 100 / 200 \\
Evaluation interval & 10 steps \\
Responses per eval problem & 16 \\
\bottomrule
\end{tabular}
\end{minipage}

\hfill
\begin{minipage}[t]{0.49\linewidth}
\captionof{table}{Method-specific math hyperparameters. \(N_1\) and \(N\)
count valid response tokens in the first and current actor batches.}
\label{tab:method-hparams}
\centering
\normalsize
\setlength{\tabcolsep}{4pt}
\renewcommand{\arraystretch}{1.0}
\begin{tabular}{@{}C{0.58\linewidth}C{0.36\linewidth}@{}}
\toprule
Hyperparameter & Value \\
\midrule
Distillation coefficient \(\lambda\) (RAD / baselines) & 0.1 \\
RAD: meta prompts & 64 \\
RAD: initial token weight \(c\) & 0 \\
RAD: weight learning rate \(\gamma\) & \(10^4 N_1/N\) \\
TIP retention ratio & 0.5 \\
RLAD \(\alpha\) & 0.5 \\
RLAD teacher log-ratio clamp & 1.0 \\
RLAD clipping thresholds & 0.20 / 0.28 \\
ExOPD \(\lambda\) & 1.25 \\
\bottomrule
\end{tabular}
\end{minipage}

\end{center}

\subsection{Initial Token Weight Ablation}
\label{app:initial-token-weights}
\begin{figure}[!b]
\centering
\includegraphics[width=0.60\linewidth]{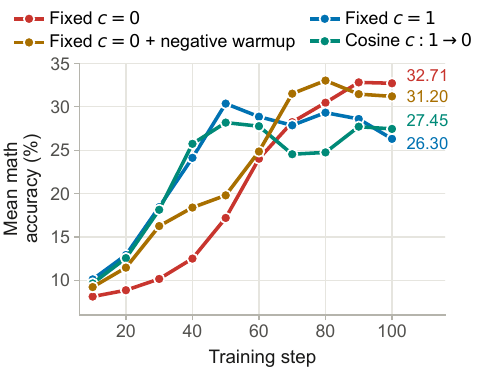}
\caption{Initial token weight and negative-weight warmup comparison for Qwen3-1.7B (non-thinking).
Each curve shows one run with seed 42 over the shared step-10--100 window.
Accuracy equally averages AIME24/25 and HMMT2025-Feb/Nov, with 16 responses
per problem. Fixed \(c=0\) without warmup uses the same run as
\cref{tab:main-results}, restricted to these four benchmarks. All runs use
AdamW lookahead, separate meta rollouts, weight learning rate
\(\gamma=10^4 N_1/N\), and no further weight normalization.
The initialization variants clip
to \([c-1,c+1]\); cosine initialization reaches \(c=0\) at step 100.
Negative-weight warmup keeps \(c=0\) and the upper bound at 1 while lowering
the lower bound from 0 to \(-1\) by step 50 using a cosine schedule.
Dots are observed evaluations; labels show step-100 accuracy.}
\label{fig:initial-token-weight-ablation}
\end{figure}

\noindent\textbf{How does the initial token weight \(c\) affect learning dynamics and final accuracy?}\enspace
In a four-run comparison, we study fixed \(c=0\), fixed \(c=1\),
a cosine schedule from \(c=1\) to \(c=0\) over 100 updates, and fixed \(c=0\)
with negative-weight warmup (\cref{fig:initial-token-weight-ablation}).
The warmup keeps the upper clipping bound at 1 and lowers the lower bound
from 0 to \(-1\) with a cosine schedule through step 50, gradually permitting
negative weights. All four runs use weight learning rate
\(\gamma=10^4 N_1/N\) without further weight normalization.
At step 100, fixed \(c=0\) reaches 32.71\% mean accuracy
without warmup and 31.20\% with warmup, compared with 26.30\% for fixed
\(c=1\) and 27.45\% for the cosine initialization schedule.
Both zero-initialized variants outperform the other initialization choices
at this checkpoint; fixed \(c=0\) without warmup leads the warmup variant by
1.51 percentage points in this single-seed comparison. The first three variants use \([c-1,c+1]\), varying initialization
and clipping bounds together; warmup changes only the lower bound while
keeping \(c=0\).

\FloatBarrier
\section{Extended Related Work}
\label{app:extended-related-work}
\noindent\textbf{On-policy distillation for LLMs.}\enspace
Knowledge distillation transfers capabilities from a larger teacher to a smaller student~\citep{buciluǎ2006model,hinton2015distilling}. For LLMs, teacher supervision can take the form of token-level probability distributions or generated text~\citep{kim2016sequence, sanh2020distilbertdistilledversionbert}. Off-policy distillation trains on fixed sequences, often generated by the teacher~\citep{hsieh-etal-2023-distilling, li-etal-2025-bild}. On-policy distillation instead queries the teacher on student-generated responses, exposing the student to supervision along its own trajectories and reducing the mismatch between training and inference~\citep{agarwal2024policy,gu2024minillm}. Self-distillation paradigm obtains supervision from the student itself, with privileged information such as verified solutions or environment feedback providing a stronger teacher context~\citep{shenfeld2026self,hubotter2026reinforcement,zhao2026self}.

\noindent\textbf{Unifying distillation and RLVR.}\enspace
Several methods combine dense teacher supervision with verifiable task rewards. KDRL~\citep{xu2025kdrl} jointly optimizes a GRPO reward objective and reverse-KL distillation, while RLAD~\citep{zhang2026reinforcement} incorporates teacher guidance into policy optimization through a likelihood ratio anchored to a mixture of the teacher and the previous student policy. RG-OPD~\citep{rgopd} uses agreement between verifier outcomes and teacher likelihoods to gate trajectory-level distillation. A related perspective connects OPD to KL-regularized reinforcement learning: ~\cite{yang2026learning} introduce a flexible reference model and scales the implicit reward derived from teacher probabilities, yielding the reward-extrapolation variant ExOPD. Our proposed method, RAD, uses task reward to determine how individual token-level distillation contributions should enter the student update.

\noindent\textbf{Token-selective distillation.}\enspace
TIP~\citep{xu2026tip} and DEAR~\citep{xiao2026finding} select teacher supervision using uncertainty, teacher--student divergence, and reasoning structure. \textsc{Reopold}~\citep{ko2026scalingreasoningefficientlyrelaxed} combines token log-ratio rewards with clipping and entropy-guided sampling. \citet{armandpour2026unmasking} diagnose token-level alignment between distillation and reward-improving directions in logit space and study oracle filtering of positively aligned tokens. RAD operationalizes reward-guided weighting during training: it differentiates a reward surrogate through the student optimizer and shared parameters to assign bounded, signed token weights. These weights can also reverse teacher contributions; our positive-only ablation supports the benefit of this additional flexibility. We compare RAD with representative reward-aware and token-selective baselines.

\noindent\textbf{Bilevel reweighting.}\enspace
Learning example weights from a held-out objective is a standard use of
bilevel optimization \citep{ren2018learning}. Gradient similarity has also
been used to adapt auxiliary losses toward a primary objective
\citep{du2018adapting}. We apply this principle at token granularity in
on-policy language-model distillation. The one-step approximation avoids
solving the full inner problem or an implicit linear system
\citep{liu2022bome,kwon2023fully}.

\section{Additional Coding Results}
\label{app:additional-coding}

\subsection{Detailed LiveCodeBench Results}
\label{app:lcb-results}

\Cref{tab:coding-results} gives the full results underlying
\cref{fig:coding-results-bars}, including standard deviations and gains
for Qwen3-1.7B (non-thinking) and OLMo3-7B-Instruct.

\begin{table}[htbp]
\caption{Full LiveCodeBench-v6 pass@1 (\%; mean $\pm$ standard deviation over four
sample positions), comparing RAD at step 100 with baselines at step 200.
Bold/underline: best/second-best trained means;
parentheses: gains of RAD in each column over the best evaluated baseline
(pp, before rounding).}
\label{tab:coding-results}
\centering
\normalsize
\setlength{\tabcolsep}{3pt}
\renewcommand{\arraystretch}{1.2}
\begin{tabularx}{\linewidth}{@{\hspace{\tabcolsep}}l*{4}{>{\centering\arraybackslash}X}@{\hspace{\tabcolsep}}}
\toprule
\radhead \multicolumn{1}{c}{Method} & Easy (322) & Medium (383) & Hard (350) & Overall (1,055) \\
\midrule
\radhead\multicolumn{5}{l}{\textit{\textbf{Student: Qwen3-1.7B (non-thinking)}}} \\
\addlinespace[1pt]
Teacher & \meanstd{96.27}{0.79} & \meanstd{69.71}{1.21} & \meanstd{26.07}{0.24} & \meanstd{63.34}{0.38} \\
\midrule
Base (student) & \meanstd{60.02}{1.17} & \meanstd{16.45}{0.26} & \meanstd{1.79}{0.71} & \meanstd{24.88}{0.35} \\
\addlinespace[1pt]
GRPO & \meanstd{66.07}{2.64} & \meanstd{23.11}{1.13} & \meanstd{4.07}{0.47} & \meanstd{29.91}{1.14} \\
OPD & \meanstd{81.37}{1.54} & \meanstd{41.38}{1.40} & \meanstd{10.79}{1.02} & \meanstd{43.44}{0.70} \\
TIP & \meanstd{82.69}{1.08} & \meanstd{40.34}{2.16} & \meanstd{9.71}{0.90} & \meanstd{43.10}{0.52} \\
RLAD & \secondmeanstd{85.56}{1.55} & \secondmeanstd{44.58}{1.24} & \secondmeanstd{11.07}{0.62} & \secondmeanstd{45.97}{0.26} \\
ExOPD & \meanstd{78.88}{0.58} & \meanstd{32.96}{1.93} & \meanstd{6.29}{0.70} & \meanstd{38.13}{0.52} \\
\midrule
\radrow\radlabel{\method{} (ours)} & \bestmeanstd{88.90}{0.40} & \bestmeanstd{57.11}{1.26} & \bestmeanstd{20.57}{0.35} & \bestmeanstd{54.69}{0.31} \\
\radgainrow\baselinegainlabel & \radgain{+3.34} & \radgain{+12.53} & \radgain{+9.50} & \radgain{+8.72} \\
\midrule
\radhead\multicolumn{5}{l}{\textit{\textbf{Student: OLMo3-7B-Instruct}}} \\
\addlinespace[1pt]
Teacher & \meanstd{89.36}{1.41} & \meanstd{64.23}{1.86} & \meanstd{28.86}{0.97} & \meanstd{60.17}{0.98} \\
\midrule
Base (student) & \meanstd{76.09}{1.58} & \meanstd{43.41}{2.31} & \meanstd{14.86}{1.73} & \meanstd{43.91}{1.50} \\
\addlinespace[1pt]
GRPO & \meanstd{84.55}{0.89} & \secondmeanstd{54.11}{1.49} & \secondmeanstd{20.57}{0.73} & \secondmeanstd{52.27}{0.26} \\
OPD & \meanstd{69.25}{3.10} & \meanstd{48.69}{1.17} & \meanstd{18.43}{1.77} & \meanstd{44.93}{1.36} \\
TIP & \meanstd{53.49}{0.83} & \meanstd{40.86}{1.30} & \meanstd{16.79}{1.02} & \meanstd{36.73}{0.37} \\
RLAD & \secondmeanstd{84.70}{2.30} & \meanstd{53.92}{0.54} & \meanstd{16.57}{0.70} & \meanstd{50.92}{0.91} \\
\midrule
\radrow\radlabel{\method{} (ours)} & \bestmeanstd{89.67}{1.86} & \bestmeanstd{63.51}{0.34} & \bestmeanstd{24.86}{1.50} & \bestmeanstd{58.67}{0.52} \\
\radgainrow\baselinegainlabel & \radgain{+4.97} & \radgain{+9.40} & \radgain{+4.29} & \radgain{+6.40} \\
\bottomrule
\end{tabularx}
\end{table}

\subsection{Qwen Coding Validation}
\label{app:coding-validation}

We evaluate Qwen3-1.7B (non-thinking), trained with
Qwen3-30B-A3B-Instruct-2507 as teacher, on the full 1,024-problem Eurus
validation split (\cref{tab:dataset-statistics}). We sample one response
per problem and report problem-weighted pass@1. Baselines generate 512
training responses per step; RAD generates 1,024 including meta rollouts.
Training and weight settings appear in Appendix~\ref{app:training-settings}.

\begin{figure}[htbp]
\centering
\includegraphics[width=\linewidth]{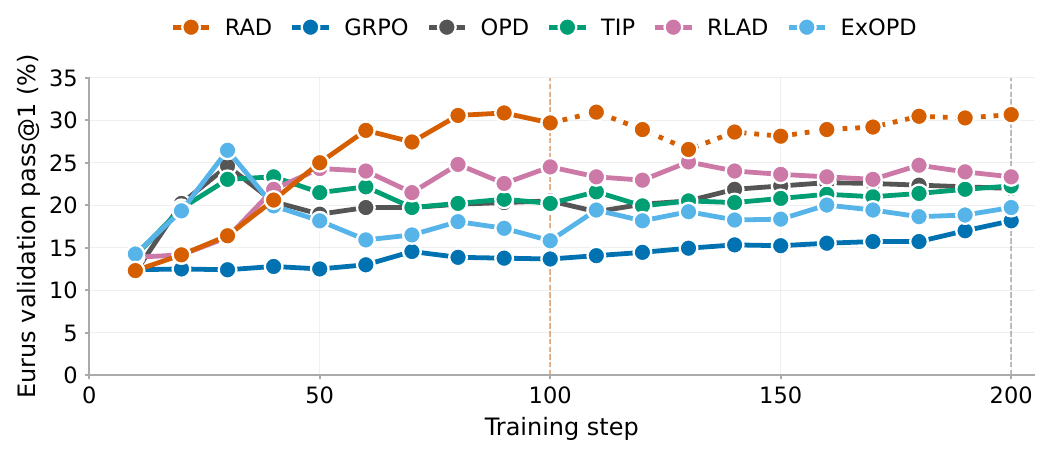}
\caption{Coding-validation dynamics for Qwen3-1.7B (non-thinking). Each point reports
pass@1 over all 1,024 validation problems for one training run per method.
Curves stop at the last available evaluation. We compare RAD at step 100
with baselines at step 200; the RAD curve is dotted beyond step 100 and
extends to step 200.}
\label{fig:coding-validation-dynamics}
\end{figure}

\begin{table}[htbp]
\caption{Qwen3-1.7B (non-thinking) coding-validation pass@1 (\%). Step-100 scores are
shown by dataset; available step-200 scores are shown overall. Overall
weights all 1,024 problems equally. Bold/underline: best/second-best
reported value per column, with ties sharing a rank; dashes: unavailable.
Parentheses give RAD's gain over the best baseline at the same step (pp,
before rounding). RAD at step 100 matches baselines at step 200 in training
responses.}
\label{tab:coding-validation-results}
\centering
\normalsize
\setlength{\tabcolsep}{3pt}
\renewcommand{\arraystretch}{1.0}
\begin{tabular*}{\linewidth}{@{\extracolsep{\fill}}lcccccc@{}}
\toprule
& \multicolumn{5}{c}{Step 100} & Step 200 \\
\cmidrule(lr){2-6}\cmidrule(l){7-7}
Method & APPS & CodeContests & Codeforces & TACO & Overall & Overall \\
\midrule
GRPO & 17.61 & 17.51 & 15.45 & 7.85 & 13.67 & 18.16 \\
OPD & 28.87 & 25.99 & 21.95 & 11.52 & 20.51 & 21.97 \\
TIP & 27.46 & 23.08 & \underline{26.83} & 12.57 & 20.21 & 22.27 \\
RLAD & \underline{37.32} & \underline{26.79} & \textbf{31.71} & \underline{15.18} & \underline{24.51} & \underline{23.34} \\
ExOPD & 19.01 & 20.16 & 16.26 & 10.21 & 15.82 & 19.73 \\
\midrule
\textbf{\method{} (ours)} & \textbf{45.07} & \textbf{32.89} & \textbf{31.71} & \textbf{20.16} & \textbf{29.69} & \textbf{30.66} \\
\baselinegainlabel & \baselinegain{+7.75} & \baselinegain{+6.10} & \baselinegain{+0.00} & \baselinegain{+4.97} & \baselinegain{+5.18} & \baselinegain{+7.32} \\
\bottomrule
\end{tabular*}
\end{table}

At the matched training-response budget, RAD at step 100 reaches 29.69\%,
exceeding RLAD at step 200 (23.34\%) by 6.35 points
(\cref{fig:coding-validation-dynamics,tab:coding-validation-results}).
Extended RAD training reaches 30.66\% at step 200.

\subsection{OLMo3-7B Settings and Validation}
\label{app:olmo-coding}

We train OLMo3-7B-Instruct with OLMo3.1-32B-Instruct as the
teacher\footnote{Model checkpoints: \url{https://huggingface.co/allenai/Olmo-3-7B-Instruct}
and \url{https://huggingface.co/allenai/Olmo-3.1-32B-Instruct}.}
on the Eurus coding split and evaluate on the same 1,024-problem validation
split described in Appendix~\ref{app:coding-validation}. We sample one
response per problem with temperature 1, top-\(p\) 1, and a 32,768-token
limit, and report problem-weighted pass@1. Each method uses one training
seed, 64 actor prompts with eight responses each, and AdamW with learning
rate \(10^{-6}\). RAD additionally uses 64 meta prompts with eight
responses each. We allocate six GPUs to student training and, for
distillation methods, two additional GPUs to teacher inference.

RAD uses the same weight settings as the Qwen coding experiment:
fixed initial token weight \(c=0\), upper bound 1, a cosine lower-bound schedule from 0 to
\(-1\) through step 50, and weight learning rate
\(\gamma=10^4 N_1/N\) without further normalization. OPD, TIP, and RAD use
distillation coefficient \(\lambda=0.1\); TIP uses full-vocabulary
reverse KL, and RLAD uses \(\alpha=0.5\).

\begin{table}[htbp]
\caption{OLMo3-7B-Instruct coding-validation pass@1 (\%) over 1,024 Eurus
problems, with one response per problem and one training run per method.
Bold/underline: best/second-best reported value per column, with ties sharing
a rank; dashes: unavailable. Parentheses give RAD's step-100 gain over the
best available baseline at the indicated step (percentage points, before
rounding).}
\label{tab:olmo-coding-results}
\centering
\normalsize
\setlength{\tabcolsep}{9pt}
\renewcommand{\arraystretch}{1.0}
\begin{tabular}{@{}lcc@{}}
\toprule
Method & Step 100 & Step 200 \\
\midrule
GRPO & \underline{34.28} & \textbf{35.84} \\
OPD & 33.20 & 33.01 \\
TIP & 27.44 & 25.68 \\
RLAD & \underline{34.28} & \underline{33.98} \\
\midrule
\textbf{\method{} (ours)} & \textbf{40.23} & --- \\
\baselinegainlabel & \baselinegain{+5.96} & \baselinegain{+4.39} \\
\bottomrule
\end{tabular}
\end{table}

\begin{figure}[htbp]
\centering
\includegraphics[width=\linewidth]{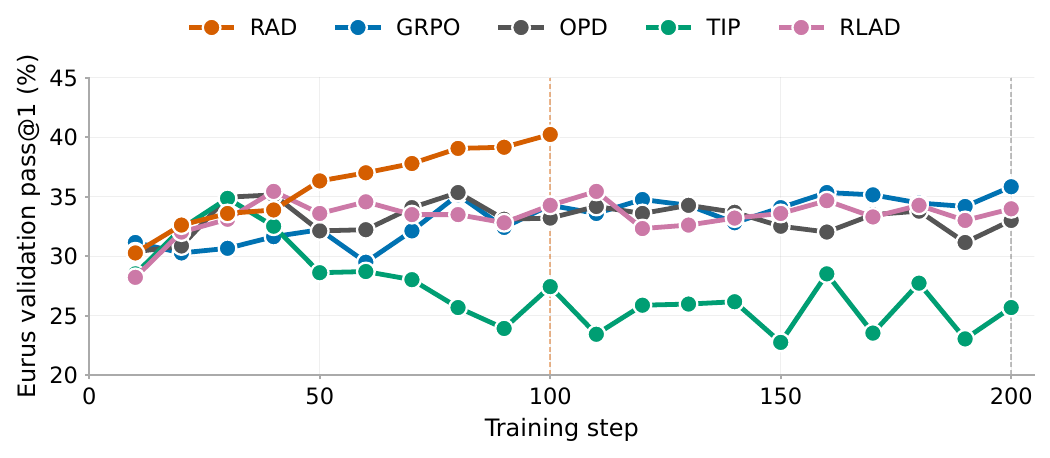}
\caption{Coding-validation dynamics for OLMo3-7B-Instruct. Each point reports
pass@1 over all 1,024 Eurus validation problems. Curves stop at the last
available evaluation. We compare RAD at step 100 with all four baselines at step 200.}
\label{fig:olmo-coding-validation-dynamics}
\end{figure}

RAD solves 412 of 1,024 validation problems at step 100, compared with 351
for both GRPO and RLAD at step 100 and 367 for GRPO at step 200
(\cref{tab:olmo-coding-results,fig:olmo-coding-validation-dynamics}).
At step 200, OPD, TIP, and RLAD reach 33.01\%, 25.68\%, and 33.98\%,
respectively. These single-run results support applying RAD
beyond the Qwen student--teacher pairs; they do not isolate the effects of
model family and student size.

We also evaluate these checkpoints on all 1,055 LiveCodeBench-v6 tasks
using the sampling and native grading protocol of the main coding
experiments, with each OLMo checkpoint's native tokenizer and chat template.
\Cref{tab:coding-results} reports the full results, including the base
student and teacher references.

\section{Preliminary Experiments on Agentic RL Settings}
\label{app:agentic}

We apply RAD to an agentic software-engineering task. A Qwen3.5-4B~\footnote{\url{https://huggingface.co/Qwen/Qwen3.5-4B}} policy is
trained on R2E-Gym~\citep{jain2025r2egym} with the OpenHands agent harness
provided by R2E-Gym. Each episode presents a repository at a failing
commit and lets the policy read, edit, and run code for up to 75 tool turns
within a 65,536-token context, generating at most 8,192 tokens per turn with
temperature 1, top-\(p\) 1, and no top-\(k\) truncation; the reward is the hidden test suite's verdict on
the final patch. Each update samples 32 tasks with eight attempts each and uses
a constant learning rate of \(10^{-6}\). The teacher is
Qwen3.6-27B.\footnote{\url{https://huggingface.co/Qwen/Qwen3.6-27B}.} The baselines train for one epoch on
R2E-Gym, and RAD for half an epoch.

We evaluate on SWE-bench Verified~\citep{jimenez2024swebench,chowdhury2024swebenchverified}
with three rollouts for each of its 500 tasks
under the training scaffold, sampling parameters, and context budget. Its
repositories are disjoint from the training corpus. pass@1 is the mean resolved
rate over all 1,500 attempts, and pass@3 is the fraction of tasks resolved by at
least one of the three attempts.

\begin{figure}[htbp]
\centering
\includegraphics[width=0.75\linewidth]{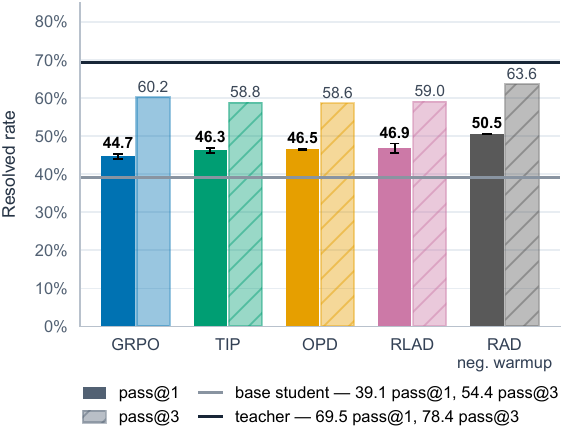}
\caption{SWE-bench Verified at a 65,536-token context and 75 tool turns. Solid
bars show pass@1 with \(\pm1\) standard deviation across the three rollouts;
hatched bars show pass@3. Horizontal lines mark the pass@1 of the base
student and the teacher.}
\label{fig:swe-verified}
\end{figure}

RAD with negative-weight warmup reaches 50.5\% pass@1 and 63.6\% pass@3, the best
of the five methods (\cref{fig:swe-verified}). It improves on GRPO by 5.9 points and
on the strongest non-RAD baseline, RLAD, by 3.7 points, while training for half
as many epochs.

\section{Computational Cost and Efficiency}
\label{app:training-time}

\subsection{Comparison with Baselines}

\Cref{tab:training-step-comparison} compares total training time for
Qwen3-1.7B (non-thinking) on math and OLMo3-7B-Instruct on coding.
RAD uses 100 updates and the baselines use 200, matching 102,400 generated
training responses. The math comparison also includes RAD-Simple at 200 updates.
We include checkpoint
saves and exclude periodic validation, process startup, and time between
training attempts. Each retained update is counted once.
For Qwen3-1.7B (non-thinking) on math, RAD-Simple records 32.48 training
hours over 200 updates, compared with 16.66 hours for RAD over 100 updates.
The main runs use six H100 GPUs for the student, whereas OPD$^2$ uses four
student GPUs alongside its two-teacher setup. These are measured costs of
the experimental configurations: generated response lengths, teacher
execution, and GPU allocations also affect step time.

\begin{table}[H]
\caption{Total training time (hours) for RAD at step 100 versus baselines at
step 200, matching generated training responses. RAD-Simple is included only
for math, at step 200. Checkpoint saves
are included; periodic validation, startup, and restart downtime are
excluded.}
\label{tab:training-step-comparison}
\centering
\normalsize
\setlength{\tabcolsep}{4pt}
\renewcommand{\arraystretch}{1.08}
\begin{tabular*}{\linewidth}{@{\extracolsep{\fill}}lccrr@{}}
\toprule
Method & Steps & \shortstack{Student\\GPUs}
& \shortstack{Qwen3-1.7B\\(non-thinking), math}
& \shortstack{OLMo3-7B-Instruct\\Coding} \\
\midrule
OPD & 200 & 6 & 14.28 & 136.51 \\
TIP & 200 & 6 & 34.09 & 166.79 \\
RLAD & 200 & 6 & 14.11 & 123.81 \\
ExOPD & 200 & 6 & 51.02 & --- \\
OPD$^2$ & 200 & 4 & 33.99 & --- \\
\midrule
RAD & 100 & 6 & 16.66 & 101.61 \\
RAD-Simple & 200 & 6 & 32.48 & --- \\
\bottomrule
\end{tabular*}
\end{table}

\subsection{RAD Stage Breakdown}

We profile the first 100 RAD updates for Qwen3-1.7B (non-thinking) on math, which uses eight NVIDIA H100 80GB GPUs: six for the student and its rollouts,
and two for teacher inference. Each step generates 512 actor responses and
512 meta responses, with a maximum generation length of 32,768 tokens.
\Cref{tab:rad-training-time} averages the recorded trainer timers over
all 100 steps, including checkpoint saves but excluding periodic validation,
job startup, and I/O outside the training-step timer.

\begin{table}[H]
\caption{RAD math training-time breakdown for Qwen3-1.7B (non-thinking),
averaged over the same 100 updates as \cref{tab:training-step-comparison}.
Entries give seconds per training step and each stage's share of recorded
training-step time. Checkpoint saves are amortized over all steps;
validation is excluded and reported separately in the text.}
\label{tab:rad-training-time}
\centering
\normalsize
\setlength{\tabcolsep}{4pt}
\renewcommand{\arraystretch}{1.08}
\begin{tabular*}{\linewidth}{@{\extracolsep{\fill}}>{\raggedright\arraybackslash}p{0.60\linewidth}rr@{}}
\toprule
Stage & Seconds & Share \\
\midrule
Distillation rollout pipeline & 211.1 & 35.2\% \\
Student log-probabilities and advantages & 11.9 & 2.0\% \\
Lookahead stage & 283.2 & 47.2\% \\
Weight computation and student update & 87.5 & 14.6\% \\
Rollout weight synchronization & 5.3 & 0.9\% \\
Checkpoint saves & 0.6 & 0.1\% \\
Other in-step overhead & 0.3 & 0.1\% \\
\midrule
\textbf{Total training step} & \textbf{599.9} & 100\% \\
\bottomrule
\end{tabular*}
\end{table}

The distillation rollout pipeline includes generation and overlapping teacher
scoring and reward evaluation. The lookahead stage includes the temporary
optimizer update, transfer of temporary weights, meta rollouts, reward-gradient
computation, and restoration of the student. Weight computation and the
committed student update include the exact JVP and projected weight update.
These are sequential trainer stages; nested teacher-request and JVP timers
are not added again. The recorded timers do not separately resolve the
components within the lookahead stage.

Mean training-step time is 10.0 minutes for Qwen3-1.7B (non-thinking),
corresponding to 16.66 recorded training hours over 100 steps.
Periodic validation adds 5.78 hours outside this total. The lookahead
stage accounts for 47.2\% of training-step time; weight
computation and the student update account for 14.6\%.
These averages include changing response lengths.

\subsection{Math Comparison at Comparable Training Times}
\label{app:math-training-time}

We use the cumulative recorded training time of RAD at step 90 for
Qwen3-1.7B (non-thinking) as a 12.09-hour reference budget.
We use TIP at step 70 and ExOPD at step 50, the first evaluated checkpoints
above this budget, for a closer time match. They use 100.9\% and 102.6\%
of the reference budget, respectively. The other baselines use their latest
available evaluated checkpoint within the budget.
Selection uses time rather than accuracy; we do not interpolate
accuracy between checkpoints. Evaluations occur every ten steps and cover
the original four benchmarks, so AIME26 and OlympiadBench are excluded from this comparison.

Training time sums the logged durations of completed updates along the
retained training trajectory, including checkpoint saves. We count each
retained update once and exclude periodic validation, process startup,
restart downtime, and discarded attempts. ExOPD logs checkpoint saves
outside its step timer, so their durations are added separately. All runs
use six H100 GPUs for the student; teacher execution follows each method's
implementation. This compares realized training times, which also depend
on generated response lengths and implementation details.

\begin{table}[H]
\caption{Math accuracy for Qwen3-1.7B (non-thinking) at comparable training
times, relative to RAD at step 90. RAD-Simple uses its latest evaluated
checkpoint within this budget.
Accuracy is the mean over AIME24/25 and HMMT2025-Feb/Nov, with standard deviation
across 16 sample-position accuracies. Budget used is the fraction of RAD's
cumulative training time. Bold/underline mark the best/second-best
means; parentheses give RAD's gain over the strongest displayed baseline (pp).}
\label{tab:math-wallclock-matched-results}
\centering
\normalsize
\setlength{\tabcolsep}{5pt}
\renewcommand{\arraystretch}{1.05}
\begin{tabular*}{\linewidth}{@{\extracolsep{\fill}}lrrrr@{}}
\toprule
Method & Step & Training (h) & Budget used & Accuracy (\%) \\
\midrule
\multicolumn{5}{l}{\textit{\textbf{Student: Qwen3-1.7B (non-thinking)}; budget 12.09 h}} \\
\addlinespace[2pt]
GRPO & 290 & 11.98 & 99.1\% & \meanstd{26.98}{2.22} \\
OPD & 160 & 11.72 & 96.9\% & \meanstd{23.23}{2.20} \\
TIP & 70 & 12.19 & 100.9\% & \meanstd{24.27}{2.62} \\
RLAD & 170 & 12.08 & 99.9\% & \meanstd{25.10}{2.67} \\
ExOPD & 50 & 12.40 & 102.6\% & \meanstd{27.34}{2.70} \\
\midrule
\textbf{\method{} (ours)} & 90 & 12.09 & 100.0\% & \secondmeanstd{32.81}{2.84} \\
RAD-Simple & 140 & 8.97 & 74.2\% & \bestmeanstd{34.64}{2.57} \\
\(\Delta\) RAD vs.\ best & --- & --- & --- & \baselinegain{+5.47} \\
\bottomrule
\end{tabular*}
\end{table}

For Qwen3-1.7B (non-thinking), RAD exceeds the strongest selected baseline,
ExOPD, by 5.47 percentage points, despite ExOPD using 2.6\% more training
time. GRPO's step 290 uses 99.1\% of the reference budget.

\subsection{Extended GRPO Training}
\label{app:extended-grpo}

\begin{figure}[H]
\centering
\includegraphics[width=\linewidth]{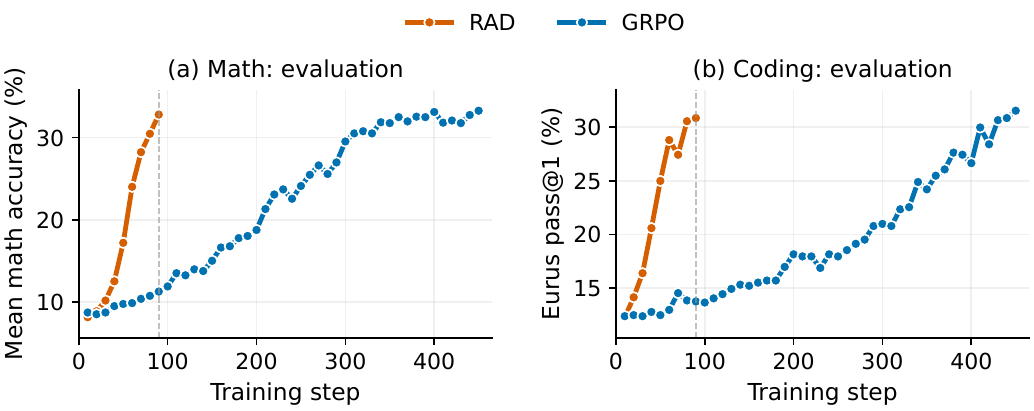}
\caption{Evaluation during extended training for Qwen3-1.7B (non-thinking).
Left: math; right: coding. Math averages AIME24/25 and HMMT2025-Feb/Nov with
16 responses per problem. Coding reports pass@1 on all 1,024 Eurus
validation problems. RAD is shown through step 90;
GRPO is shown through step 450 in both math and coding.
The vertical line marks step 90.}
\label{fig:rad100-grpo-extended}
\end{figure}

\Cref{fig:rad100-grpo-extended} compares Qwen3-1.7B (non-thinking) RAD runs
through step 90 with GRPO over a 450-step window. RAD reaches 32.81\%
on the four math benchmarks and 30.86\% on Eurus coding validation at
step 90. At step 450, GRPO reaches
33.28\% on math and 31.54\% on coding.
GRPO generates 512 training responses per step, while RAD generates
512 actor and 512 meta responses. Both GRPO endpoints use 2.50
times as many training responses as RAD at step 90.

\section{Gradient of Weighted On-Policy Distillation}
\label{app:weighted-gradient}

\noindent\textbf{Per-token surrogate.}\enspace
For a response sampled from the frozen policy \(\pi_{\mathrm{old}}\), the
surrogate used in \cref{eq:weighted-rkl,eq:batch-rkl} is
\begin{equation}
    \ell_{D,t}(\btheta;\by)
    =\rho_t(\btheta)
    \log\frac{\pi_{\btheta}(y_t\mid s_t)}
              {\pi_{\mathrm T}(y_t\mid s_t)},
    \qquad
    \rho_t(\btheta)
    =\frac{\pi_{\btheta}(y_t\mid s_t)}
           {\pi_{\mathrm{old}}(y_t\mid s_t)}.
    \label{eq:token-surrogate}
\end{equation}
The sampling policy and prefixes are held fixed. For batch notation, write
\(\ell_{D,i,t}(\btheta;\by_i)=\ell_{D,t}(\btheta;\by_i)\).

\noindent\textbf{Derivation of the per-token gradient.}\enspace
At a fixed prefix \(s_t\), consider the expected weighted reverse KL
\[
    L_{D,t}(\btheta,w_t\mid s_t)
    =\mathbb E_{y_t\sim\pi_{\btheta}(\cdot\mid s_t)}
    \left[
        w_t\log\frac{\pi_{\btheta}(y_t\mid s_t)}
                      {\pi_{\mathrm T}(y_t\mid s_t)}
    \right].
\]
Holding \(w_t\) fixed when differentiating with respect to \(\btheta\) gives
\begin{align}
    \nabla_{\btheta}L_{D,t}(\btheta,w_t\mid s_t)
    &=
    \nabla_{\btheta}
    \sum_{y_t}
        \pi_{\btheta}(y_t\mid s_t)w_t
        \log
        \frac{\pi_{\btheta}(y_t\mid s_t)}
             {\pi_{\mathrm{T}}(y_t\mid s_t)}
    \nonumber\\
    &=
    \mathbb E_{y_t}\left[
        w_t
        \left(
            \log
            \frac{\pi_{\btheta}(y_t\mid s_t)}
                 {\pi_{\mathrm{T}}(y_t\mid s_t)}
            +1
        \right)
        \nabla_{\btheta}\log\pi_{\btheta}(y_t\mid s_t)
    \right] \nonumber\\
    &=
    \mathbb E_{y_t}\left[
        w_t
        \log
        \frac{\pi_{\btheta}(y_t\mid s_t)}
             {\pi_{\mathrm{T}}(y_t\mid s_t)}
        \nabla_{\btheta}\log\pi_{\btheta}(y_t\mid s_t)
    \right]
    \nonumber\\
    &\qquad+
    \mathbb E_{y_t}\left[
        w_t
        \nabla_{\btheta}\log\pi_{\btheta}(y_t\mid s_t)
    \right].
    \label{eq:appendix-local-gradient}
\end{align}
This is the per-token, frozen-prefix gradient used in \cref{sec:method}.
The surrogate in \cref{eq:token-surrogate} realizes the same gradient.
At the update point \(\pi_{\btheta}=\pi_{\mathrm{old}}\), the ratio
\(\rho_t=1\) in value but remains differentiable, giving
\[
    \nabla_{\btheta}
    \ell_{D,i,t}(\btheta;\by_i)
    =
    \left(
        \log
        \frac{\pi_{\btheta}(y_{i,t}\mid s_{i,t})}
             {\pi_{\mathrm{T}}(y_{i,t}\mid s_{i,t})}
        +1
    \right)
    \nabla_{\btheta}\log\pi_{\btheta}(y_{i,t}\mid s_{i,t}).
\]
Since the sampling distribution in \cref{eq:weighted-rkl} is fixed, its
gradient is simply
\[
    \nabla_{\btheta}L_D(\btheta,\bw)
    =\mathbb E_{\by\sim\pi_{\mathrm{old}}(\cdot\mid\bx)}
    \left[
        \sum_t w_t(\by)\nabla_{\btheta}\ell_{D,t}(\btheta;\by)
    \right].
\]
Student rollouts provide a sampled estimate of this expected gradient.

\noindent\textbf{Action-independent weights and OPD.}\enspace
If \(w_t\) does not depend on \(y_t\), the second expectation in
\cref{eq:appendix-local-gradient} vanishes by the zero-mean score identity.
The result is \(w_t\) times the OPD gradient, with OPD recovered at
\(w_t=1\). For an action- or response-dependent weight, this term
generally does not vanish. If the weight had explicit differentiable
dependence on \(\btheta\), an additional term
\[
    \mathbb E_{y_t}\left[
        \log
        \frac{\pi_{\btheta}(y_t\mid s_t)}
             {\pi_{\mathrm{T}}(y_t\mid s_t)}
        \nabla_{\btheta}w_t
    \right]
\]
would appear;
the updated weights in \cref{eq:batch-rkl} are instead detached during
student training, and its normalization is fixed.

\noindent\textbf{Full sequence-level gradient.}\enspace
Equation~\eqref{eq:appendix-local-gradient} conditions on a fixed prefix and is
therefore a per-token gradient. For comparison, define the full sequence-level
objective
\[
    L_D^{\mathrm{seq}}(\btheta,\bw)
    =\mathbb E_{\by\sim\pi_{\btheta}(\cdot\mid\bx)}
    \left[
        \sum_t w_t(\by)
        \log\frac{\pi_{\btheta}(y_t\mid s_t)}
                  {\pi_{\mathrm T}(y_t\mid s_t)}
    \right].
\]
Unlike \cref{eq:weighted-rkl}, its sampling distribution depends on
\(\btheta\). Differentiating the complete trajectory expectation gives
\begin{align*}
    \nabla_{\btheta}
    L_D^{\mathrm{seq}}(\btheta,\bw)
    &=
    \mathbb E_{\by}\left[
        \sum_t w_t(\by)
        \nabla_{\btheta}
        \log
        \frac{\pi_{\btheta}(y_t\mid s_t)}
             {\pi_{\mathrm{T}}(y_t\mid s_t)}
    \right]
    \\
    &\quad+
    \mathbb E_{\by}\left[
        \left(
            \sum_t w_t(\by)
            \log
            \frac{\pi_{\btheta}(y_t\mid s_t)}
                 {\pi_{\mathrm{T}}(y_t\mid s_t)}
        \right)
        \sum_k\nabla_{\btheta}\log\pi_{\btheta}(y_k\mid s_k)
    \right].
\end{align*}
The first term supplies the \(+1\) contribution because the teacher is fixed
and
\[
\nabla_{\btheta}
\log
\frac{\pi_{\btheta}(y_t\mid s_t)}
     {\pi_{\mathrm{T}}(y_t\mid s_t)}
=\nabla_{\btheta}\log\pi_{\btheta}(y_t\mid s_t).
\]
Combining this with the
diagonal \(k=t\) contribution after expanding the second term gives
\[
    w_t(\by)
    \left(
        \log
        \frac{\pi_{\btheta}(y_t\mid s_t)}
             {\pi_{\mathrm{T}}(y_t\mid s_t)}
        +1
    \right)
    \nabla_{\btheta}\log\pi_{\btheta}(y_t\mid s_t),
\]
which is the frozen-prefix contribution used by the main derivation. The
remaining off-diagonal \(k\ne t\) terms are the cross-token
trajectory-distribution gradients omitted by the fixed-prefix estimator.
Our implementation uses this fixed-prefix estimator and refreshes the state
distribution with new on-policy rollouts at every iteration.

\section{Hypergradient through One Optimizer Step}
\label{app:optimizer-hypergradient}

For the fixed batch in \cref{eq:batch-rkl}, let
\(\boldsymbol g(\bw)=\nabla_{\btheta}\widehat L_D(\btheta,\bw;\mathcal B)\).
One optimizer step and its evaluation at initial token weight \(c\) are
\[
    \btheta^+(\bw)=\mathcal A(\btheta,\boldsymbol g(\bw)),
    \qquad \btheta^+=\btheta^+(c\mathbf 1).
\]
The batch, original parameters, and current optimizer state are fixed.
Using the reward direction \(\boldsymbol p\) from
\cref{eq:optimizer-reward-direction}, the weight derivative is
\begin{equation}
    \left.\frac{\partial J_R(\btheta^+(\bw))}{\partial w_{i,t}}
    \right|_{\bw=c\mathbf 1}
    =-\frac{\lambda}{N}\left\langle
        \boldsymbol p,\nabla_{\btheta}\ell_{D,i,t}(\btheta;\by_i)
    \right\rangle
    =\frac{\lambda}{N}I_{i,t}.
    \label{eq:general-optimizer-hypergradient}
\end{equation}
The JVP computes the scores \(I_{i,t}\) together. We apply the weight
learning rate \(\gamma\) directly to these scores in
\cref{eq:projected-weight-update}.

\noindent\textbf{SGD.}\enspace

For an unclipped SGD step without momentum,
\[
    \btheta^+
    =
    \btheta-\eta\boldsymbol{g},
    \qquad
    \boldsymbol{p}
    =
    \eta\nabla_{\btheta^+}J_R(\btheta^+).
\]
This gives the SGD score in \cref{eq:token-score}, with the learning rate
already included. Momentum or gradient clipping is handled by including
that operation in \(\mathcal A\).

\noindent\textbf{AdamW.}\enspace

Our experiments use AdamW with global gradient clipping. Let
\(\widetilde{\boldsymbol{g}}(\bw)=\mathcal C(\boldsymbol{g}(\bw))\), where \(\mathcal C\) is the
clipping operator; we suppress the dependence on \(\bw\) below. Given the current moments
\((\boldsymbol{u},\boldsymbol{v})\) and update index \(k\), AdamW computes
\begin{align*}
    \boldsymbol{u}^+
    &=
    \beta_1\boldsymbol{u}
    +(1-\beta_1)\widetilde{\boldsymbol{g}},
    &
    \boldsymbol{v}^+
    &=
    \beta_2\boldsymbol{v}
    +(1-\beta_2)\widetilde{\boldsymbol{g}}
        \odot\widetilde{\boldsymbol{g}},
    \nonumber\\
    \widehat{\boldsymbol{u}}^+
    &=
    \frac{\boldsymbol{u}^+}{1-\beta_1^k},
    &
    \widehat{\boldsymbol{v}}^+
    &=
    \frac{\boldsymbol{v}^+}{1-\beta_2^k},
    \nonumber\\
    \btheta^+(\bw)
    &=
    (1-\eta d)\btheta
    -
    \eta\,
    \widehat{\boldsymbol{u}}^+
    \oslash
    \left(\sqrt{\widehat{\boldsymbol{v}}^+}
    +\epsilon_{\mathrm{A}}\right).
\end{align*}
Here \(\eta\) is the learning rate, \(d\) is weight decay,
\(\beta_1,\beta_2\) are the moment coefficients, and
\(\epsilon_{\mathrm{A}}\) is the Adam numerical-stability constant. All
vector operations are elementwise; \(\odot\) and \(\oslash\) denote
multiplication and division.

Let \(D\) be the Jacobian of
\(\widehat{\boldsymbol{u}}^+/
(\sqrt{\widehat{\boldsymbol{v}}^+}+\epsilon_{\mathrm{A}})\) with respect to
\(\widetilde{\boldsymbol{g}}\). Specializing the optimizer-aware reward
direction to AdamW gives
\begin{equation}
    \boldsymbol{p}
    =
    \eta
    \left(
        \frac{\partial\mathcal C}{\partial\boldsymbol{g}}
    \right)^{\!\top}
    D^\top
    \nabla_{\btheta^+}J_R(\btheta^+).
    \label{eq:adamw-reward-direction}
\end{equation}
Substituting \cref{eq:adamw-reward-direction} gives the AdamW-specialized JVP
used in training:
\begin{equation}
    \bI(\by)
    =
    -
    \operatorname{JVP}_{\btheta}
    \left[
        \boldsymbol{\ell}_D(\btheta;\by)
    \right]
    \left[
        \eta
        \left(
            \frac{\partial\mathcal C}{\partial\boldsymbol{g}}
        \right)^{\!\top}
        D^\top
        \nabla_{\btheta^+}J_R(\btheta^+)
    \right].
    \label{eq:adamw-token-jvp}
\end{equation}

\section{Custom Triton Kernel for Attention JVPs}
\label{app:attention-jvp}

In the \texttt{flash\_triton} attention implementation, FlashAttention computes
the attention output and row log-sum-exp during the model JVP in
\cref{eq:token-score-jvp}, and a custom Triton rule propagates their
directional derivatives. We describe scaled dot-product self-attention
with zero dropout and no score softcapping. For one sequence and head,
\[
    S=QK^\top/\sqrt d,\qquad P=\operatorname{softmax}(S),\qquad
    O=PV,\qquad L_i=\log\sum_j\exp(S_{ij}),
\]
where \(Q,K,V\) are the queries, keys, and values, and \(d\) is the head
dimension. Softmax and log-sum-exp include only allowed keys in each row;
masked probabilities are zero. Structural masks are fixed during differentiation.
Dots denote
directional derivatives induced by the parameter direction
\(\boldsymbol p\), with \(\dot Q,\dot K,\dot V\) supplied by forward-mode
differentiation through the preceding operations. The attention JVP is
\begin{align}
    \dot S&=(\dot QK^\top+Q\dot K^\top)/\sqrt d,
    & a_i&=\sum_jP_{ij}\dot S_{ij},\nonumber\\
    \dot O&=P\dot V+(P\odot\dot S)V-a\odot O,
    & \dot L&=a.
    \label{eq:attention-jvp}
\end{align}
Here \(\odot\) is elementwise multiplication, with \(a\) broadcast across
each output row. The term \(-a\odot O\) accounts for differentiating the
softmax normalizer. Reusing the saved \(O,L\), the kernel reconstructs
probabilities one tile at a time and accumulates \cref{eq:attention-jvp}.

\begin{algorithm}[H]
\caption{Tiled attention JVP using the FlashAttention output and log-sum-exp}
\label{alg:attention-jvp}
\begin{algorithmic}[1]
\Require \(Q,K,V\), tangents \(\dot Q,\dot K,\dot V\), saved \(O,L\),
sequence boundaries, and attention masks.
\Ensure Attention-output tangent \(\dot O\) and log-sum-exp tangent \(\dot L\).
\State Initialize \(\dot O,\dot L\) to zero, including padded positions.
\For{each sequence, query head, and valid query block \(B\) in parallel}
    \State Select the corresponding key/value head for grouped-query attention.
    \State Load \(Q_B,\dot Q_B,O_B,L_B\); initialize FP32 accumulators
        \(U_B\gets0\), \(a_B\gets0\).
    \For{each key block \(C\) in the same sequence allowed by the block's masks}
        \State Load \(K_C,V_C,\dot K_C,\dot V_C\).
        \State \(S_{BC}\gets Q_BK_C^\top/\sqrt d\).
        \State \(\dot S_{BC}\gets(\dot Q_BK_C^\top+Q_B\dot K_C^\top)/\sqrt d\).
        \State \(P_{BC}\gets\exp(S_{BC}-L_B)\), broadcasting \(L_B\) over keys.
        \State Set masked entries of \(P_{BC}\) to zero.
        \State \(a_B\gets a_B+\operatorname{rowsum}(P_{BC}\odot\dot S_{BC})\).
        \State \(U_B\gets U_B+P_{BC}\dot V_C+(P_{BC}\odot\dot S_{BC})V_C\).
    \EndFor
    \State Store \(\dot O_B\gets U_B-a_B\odot O_B\) and \(\dot L_B\gets a_B\)
        for valid queries.
\EndFor
\end{algorithmic}
\end{algorithm}

Query blocks stay within sequence boundaries, so packed sequences cannot
attend to one another. Causal, padding, and sliding-window masks use the
same valid query--key pairs as the primal; grouped-query heads share their
key/value tensors without explicitly repeating them. Our training launchers
use 16-query blocks, with Triton autotuning key-block sizes of 16 or 32.
Probability factors and tangent accumulators remain in FP32; the products
with \(V\) and \(\dot V\) use \texttt{tf32x3} tensor-core precision.
The kernel stores \(\dot O\) in the primal output's dtype and \(\dot L\) in FP32.
It avoids storing full attention or attention-tangent matrices, while
full-attention arithmetic remains quadratic in sequence length. The rule
analytically differentiates the attention equations up to floating-point error.

\subsection{Layer-Streamed JVP Execution}
\label{app:layer-streamed-jvp}

The \texttt{olmo3\_layer\_streamed} backend evaluates the embedding, decoder
layers, final normalization, and output projection in sequence. For each
parameter group, the FSDP executor gathers its parameters and the corresponding
components of \(\boldsymbol p\), propagates both activations and activation
tangents, and releases the gathered tensors after use. It can prefetch the
next group, so gathered parameter and direction storage is bounded by the
current and prefetched groups. Activation states are retained for a bounded
set of microbatches at a time, with microbatch grouping synchronized across
ranks to preserve the order of collective operations.

Within each OLMo decoder layer, the backend uses the same FlashAttention
primal and Triton attention tangent as \texttt{flash\_triton}
(\cref{alg:attention-jvp}), including the masks for full and sliding-window
attention. Forward-mode differentiation propagates the chain-rule JVP through
successive layers, reducing the memory needed for gathered parameters and
their direction while retaining the token-score computation in
\cref{eq:token-score-jvp}.

\end{document}